\documentclass{article}
\usepackage[final,main]{neurips_2026}
\usepackage[utf8]{inputenc}
\usepackage[T1]{fontenc}
\usepackage{hyperref}
\usepackage{url}
\usepackage{booktabs}
\usepackage{amsfonts}
\usepackage{amsmath}
\usepackage{nicefrac}
\usepackage{microtype}
\usepackage{xcolor}
\usepackage{graphicx}
\usepackage{subcaption}
\usepackage{enumitem}
\usepackage{float}
\usepackage{wrapfig}
\usepackage{verbatim}
\usepackage{placeins}

\renewcommand{\answerTODO}[1][]{\textcolor{blue}{\bf [TODO]}}

\graphicspath{{figures/}}

\title{Tracing Psychometric Inference in Large Language Models}
\author{Chih-Hao Hsu \\
  \textit{National Taiwan University} \\
  \And
  Feng-Chun B. Chou \\
  \textit{National Taiwan University} \\
  \And
  Pin-Hao A. Chen\thanks{Corresponding author. Email: {andyphchen@ntu.edu.tw}.} \\
  \textit{National Taiwan University}}

\begin{document}
\setcounter{footnote}{1}
\maketitle
\setcounter{footnote}{0}

\begin{abstract}
Do large language models represent psychological constructs internally, or do they only generate outputs that resemble psychometric structure? We address this question using a cross-persona paradigm across 14 model variants from two families (Llama~3 and Qwen~2.5, 0.5B--14B, base and instruct). Given responses on one psychological scale from real individuals ($N = 272$ Taiwanese university students), models predict responses on six other scales. At the behavioral level, larger variants reproduce and amplify human cross-scale correlations; amplification persists after correcting the seven-scale human correlations for measurement attenuation. We then examine whether this structure is reflected internally. Contrastive direction analysis finds geometry aligned with the human matrix in larger instruct variants, but little alignment in base variants under this assay. Across models, geometry strength is associated with behavioral amplification ($r = 0.68$, $p = 0.008$; partial $r = 0.69$ controlling for log size). To relate internal structure to output behavior, a matched activation-probing paradigm shows that probe slopes vary less than behavioral slopes ($1.38$--$1.77$ vs $0.52$--$1.53$). A synthetic control shows that ridge probing can inflate these slopes; after subtracting a baseline based on a reliability proxy from another analysis, adjusted slopes remain associated with behavior ($r = 0.88$). We term the measured probe--behavior gap \emph{readout attenuation}. These results establish decodability and geometric alignment under the tested assays, without showing that the measured directions mediate the cross-persona answers.
\end{abstract}

\section{Introduction}
\label{sec:intro}

A growing body of work shows that large language models can simulate human psychological profiles at the group level, producing response patterns that resemble survey data in structure and covariance \citep{liu2026five, argyle2023out, serapio2023personality}. This raises a fundamental interpretability question beyond behavioral evaluation: when an LLM reproduces psychometric structure in its output, does it reflect internal representations of psychological constructs, or merely generate similar patterns from surface-level statistical associations? A model that emulates without internal representation is a convincing imitator, not a system that ``has'' the structure in any meaningful sense. The same ambiguity arises in LLM interpretability regarding whether next-token predictors develop world models of domains they reason about \citep{li2023emergent, nanda2023emergent}, and makes the psychometric case particularly tractable: psychology supplies a rich, quantitative reference in the $16 \times 16$ human subscale correlation matrix. Here the reference comes from one Taiwanese student sample assessed in Traditional Chinese.

We design a cross-persona paradigm to probe this question: a subject's responses on one psychological scale are placed in the LLM prompt, and the model is asked to complete items from a different, unseen scale. This paradigm has two properties that make it well-suited for probing internal representation. First, by anchoring the model to a specific human subject's response pattern rather than eliciting generic self-report, it asks the model to perform cross-construct inference on a per-subject basis, generating a subject-by-construct response matrix directly comparable to the human ground truth. Second, by rotating which scale serves as the persona, cross-scale relationships are evaluated in both directions, yielding a bidirectional 16$\times$16 correlation matrix that characterizes the between-construct structure rather than a one-way projection from a single anchor scale (e.g., Big Five used as fixed input in \citealt{liu2026five}). Behaviorally, across 14 model variants from two families (Llama~3 and Qwen~2.5, 0.5B--14B, base and instruct) and $N = 272$ participants, LLMs reproduce and systematically amplify human cross-scale correlations, with amplification persisting after correcting for measurement attenuation (\S\ref{sec:setup}). We examine this amplification in model internals using contrastive direction extraction, activation probing, and steering, asking whether (a) psychometric structure is encoded in mid-layer representations, (b) amplification varies across model depths, and (c) representational strength is associated with behavior across models.

We distinguish three levels of evidence: a probe can show that cross-scale information is linearly decodable from activations; contrastive directions can show that construct similarities are organized like the human reference; and steering tests whether changing a direction affects responses. The first two analyses do not, by themselves, establish that these directions mediate the model's answers to the cross-persona task; our steering experiments are exploratory.

These analyses yield three contributions:
\begin{enumerate}[nosep,leftmargin=*]
\item \textbf{Structured psychometric geometry is detected most clearly in larger instruct variants.} Contrastive directions for 16 subscales recover a similarity structure aligned with human psychometric correlations in these variants, but show little alignment in the tested base models (Mantel $r$ up to $0.66$ for Qwen~14B Instruct; $\leq 0.14$ for all base). Notably, base and instruct models with comparable behavioral performance can differ sharply in whether such representational structure is present under this assay.
\item \textbf{Cross-scale information is decodable from decision-point activations.} The item-level probe yields slopes $1.38$--$1.77$, versus behavioral slopes $0.52$--$1.53$. A synthetic control shows that the probe itself inflates slopes; after proxy-based subtraction, adjusted slopes remain associated with behavior ($r = 0.88$, $p = 0.002$). We use \emph{readout attenuation} as an operational name for the measured pipeline gap.
\item \textbf{Representational geometry is associated with behavioral amplification across models} ($r = 0.68$, $p = 0.008$; partial $r = 0.69$ controlling for log size), with a stronger relationship observed among instruct models ($r = 0.90$, Fisher-$z$ $p = 0.064$ given $n = 7$ per group).
\end{enumerate}

\section{Related Work}
\label{sec:related}

\paragraph{LLM psychometric simulation.}
A growing body of work examines the ability of LLMs to simulate human psychological profiles and individual agents \citep{park2024generative, aher2023using, dillion2023can, jiang2024personallm, choi2024picle}. Closest to our work, \citet{liu2026five} prompted LLMs with Big Five responses to predict nine other scales, reporting strong alignment and systematic amplification, which they attribute to noise filtering (we revisit this account in \S\ref{sec:discussion}). Related work examines LLM personality profiles \citep{serapio2023personality, sorokovikova2024llms}, psychometric benchmarks \citep{li2024evaluating, samuel2024personagym}, response biases \citep{tjuatja2024llms}, and the validity of applying psychological instruments to LLMs \citep{pellert2024ai, dong2025humanizing}. We extend this line of work by moving from behavioral characterization to a representation-level account.

\paragraph{Representation engineering and probing.}
Linear probing and contrastive methods identify interpretable directions in LLM activation space corresponding to concepts such as truthfulness \citep{marks2024geometry, burns2023discovering, li2024inference}, spatial and temporal structure \citep{gurnee2024language}, linear features more generally \citep{park2024linear, elhage2022toy}, world models in trained sequence models \citep{li2023emergent, nanda2023emergent}, and persona traits \citep{chen2025persona, zou2023representation}. Activation Addition and Contrastive Activation Addition demonstrate that adding vectors along these directions causally steers model behavior \citep{turner2024activation, zou2023representation}. We adapt these methods to psychometric constructs, using contrastive direction extraction and activation steering to probe how psychological traits are represented and processed.

Recent studies have also used probes and interventions to study affect, personality, and moral concepts in LLMs \citep{keeman2026whether,dan2026traits,yu2026moral,schacht2025moral}. Our question differs in comparing a population-derived cross-construct correlation matrix with both model behavior and activation geometry across base and instruction-tuned variants.

\section{Setup: Cross-Persona Paradigm and Behavioral Baseline}
\label{sec:setup}

\begin{figure}[t]
\centering
\includegraphics[width=0.70\linewidth]{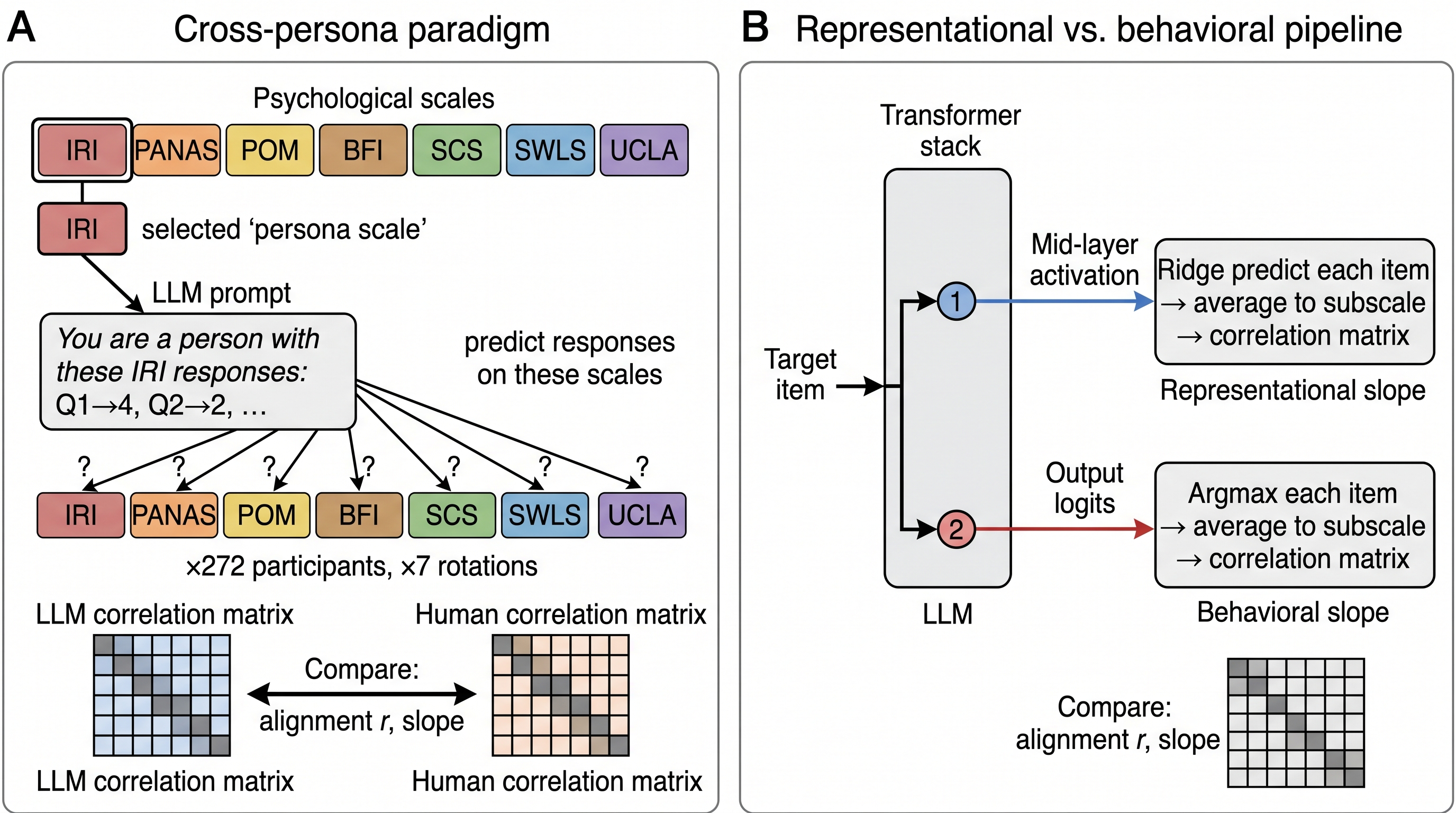}
\caption{Overview of experimental design. \textbf{Left}: Cross-persona paradigm: a subject's responses on one persona scale are provided in the prompt, and the model predicts responses on other scales across all 7 scale rotations. \textbf{Right}: Behavioral and representational pipelines extract parallel $16 \times 16$ cross-scale correlation matrices from argmax outputs and mid-layer ridge-predicted responses, respectively, and compare them to the human reference.}
\label{fig:overview}
\end{figure}

\subsection{Human psychometric data}

We use psychometric data from $N = 272$ participants who completed seven scales in Traditional Chinese, encompassing 114 items and 16 subscales (Table~\ref{tab:scales}). These scales span diverse psychological domains including empathy (IRI), affect (PANAS), well-being (POM, SWLS), personality (BFI-10), self-construal (SCS), and loneliness (UCLA). Consistent with prior cross-scale analyses of LLM psychometric behavior \citep{liu2026five}, the main human reference is the $16 \times 16$ subscale correlation matrix (120 off-diagonal pairs); scale-level analysis ($7\times 7$) yields too few pairs while item-level ($114\times 114$) suffers from within-subscale redundancy. Bootstrap resampling (2000 subject-level iterations) confirms matrix stability ($r = 0.97$, 95\% CI: $[0.96, 0.99]$; App~\ref{app:robustness}).

These responses came from a separate research project with Taiwanese university students. Ages were 17--32 years (mean 19.75, SD 1.92); 150 reported female and 122 male. Participants provided informed consent and received either course credit or monetary compensation. The human data collection was approved by the Research Ethics Center at National Taiwan University (approval no. 202206HM013). Model prompts use the Traditional Chinese item wording and response scales administered to the participants. The exact wording is included in the accompanying repository. The reference is specific to this sample and administration.

\begin{table}[t]
\centering\footnotesize
\caption{Scale inventory. PT = Perspective Taking, FS = Fantasy, EC = Empathic Concern, PD = Personal Distress, PA = Positive Affect, NA = Negative Affect, E = Extraversion, A = Agreeableness, C = Conscientiousness, N = Neuroticism, O = Openness.}
\label{tab:scales}
\begin{tabular}{llccl}
\toprule
Scale & Abbrev. & Items & Range & Subscales \\
\midrule
Interpersonal Reactivity Index \citep{davis1980multidimensional} & IRI & 28 & 1--5 & PT, FS, EC, PD \\
Positive \& Negative Affect Schedule \citep{watson1988panas} & PANAS & 20 & 1--5 & PA, NA \\
Peace of Mind \citep{lee2013peace} & POM & 7 & 1--6 & POM \\
Big Five Inventory-10 \citep{rammstedt2007measuring} & BFI & 10 & 1--7 & E, A, C, N, O \\
Self-Construal Scale \citep{singelis1994self} & SCS & 24 & 1--7 & Indep, Interdep \\
Satisfaction with Life Scale \citep{diener1985satisfaction} & SWLS & 5 & 1--7 & LifeSat \\
UCLA Loneliness Scale \citep{russell1996ucla} & UCLA & 20 & 1--4 & Loneliness \\
\bottomrule
\end{tabular}
\end{table}

\subsection{Cross-persona paradigm}

Each of the seven scales serves as the \emph{persona scale} in turn. A subject's item-level responses on the persona scale are placed in the system prompt (``You are completing psychological questionnaires as a person with the following responses on the [scale name]: Q1.~[text]~$\rightarrow$~[response] ...''), and the model is asked to respond to items from all other scales. Each target item is administered as a separate call, with the response extracted as the argmax over valid response option log-probabilities. This yields 7 rotations per model, each producing cross-scale responses for all 272 subjects. We verify the resulting $16 \times 16$ correlation matrix against trivial in-prompt copying: within-scale subscale pairs (sharing the persona scale, where direct context is available) are not systematically closer to human correlations than cross-scale pairs (App~\ref{app:trivial_context}).

We evaluate 14 model variants from two open-weight families: Qwen~2.5 (0.5B, 3B, 7B, 14B; \citealt{yang2024qwen25}) and Llama~3 (1B, 3B, 8B; \citealt{grattafiori2024llama3}), each in base and instruct variants. All models receive identical raw-text prompts (no chat template) to isolate the effect of instruction tuning weights from prompt formatting. A matched chat-template control is reported in App~\ref{app:prompt_controls}.

\subsection{Behavioral results}

For each model, we construct an implicit subscale-level cross-scale correlation matrix by correlating persona-conditioned subscale scores (mean of argmax integer item responses) with human subscale scores across subjects. We compare the between-scale entries with the human reference using Pearson correlation (alignment $r$) and OLS slope. Alignment asks whether the model preserves which relationships between subscales are stronger, weaker, positive, or negative. The slope measures their overall magnitude relative to the human matrix. A model can match the pattern while exaggerating the correlations; the two measures can also rise together across variants. Headline values average the seven persona-scale rotation estimates, rather than fitting one regression over pooled matrix entries. Matrix pairs share constructs and subjects and are not independent observations.

\begin{figure}[t]
\centering
\includegraphics[width=0.65\linewidth]{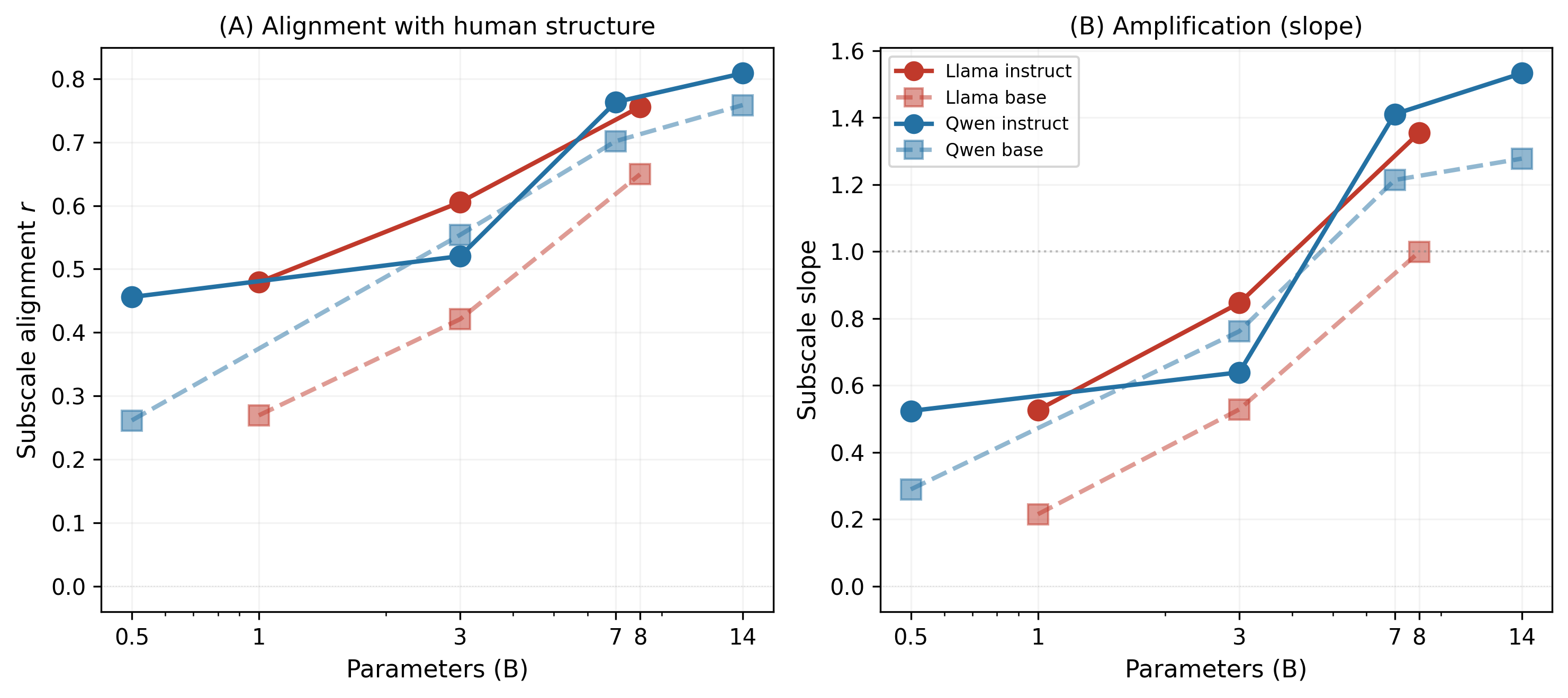}
\caption{Behavioral scaling across 14 variants at the subscale level. \textbf{Left}: mean alignment $r$ across seven persona-scale rotations. \textbf{Right}: mean uncorrected between-scale slope across the same rotations. Line style distinguishes base and instruct variants; the dotted horizontal line marks slope $=1$.}
\label{fig:behavioral}
\end{figure}

Two patterns emerge (Figure~\ref{fig:behavioral}). First, alignment increases with model size, with larger models more closely reproducing human psychometric structure. Second, the slope transitions from attenuation ($< 1$) in smaller models to amplification ($> 1$) in larger models, with instruct models showing stronger amplification at scale. Correcting the human seven-scale matrix for measurement attenuation using Cronbach's $\alpha$ \citep{nunnally1975psychometric} reduces but does not eliminate amplification in the tested $\geq 7$B models: corrected slopes remain $1.19$--$1.39$ (App~\ref{app:behavioral}). The OLS intercept is also positive across the 14 pooled 16-subscale comparisons (App~\ref{app:behavioral}). In contrast, \citet{liu2026five} report corrected slopes near $1.0$. The persistence of amplification after correction motivates our representation-level investigation.

\paragraph{Prompt and response-readout controls.}
We tested prompt and readout choices on a matched subset of 50 participants with Qwen~14B Instruct. Removing scale names gave alignment $r=0.753$ and slope $1.401$ under argmax readout; using the official chat template gave $r=0.766$ and slope $1.330$. Expected-value readout under the standard prompt gave slope $1.438$. These are single-model checks; App~\ref{app:prompt_controls} reports the remaining conditions and valid-pair counts. A human split-half baseline is reported in App~\ref{app:robustness}.

\section{Psychometric Geometry in Activation Space}
\label{sec:geometry}

\subsection{Contrastive direction extraction}

Adapting the persona vectors approach \citep{chen2025persona}, for each of the 16 subscales we extract a contrastive direction $\mathbf{d}_k = \bar{\mathbf{h}}_k^{\text{high}} - \bar{\mathbf{h}}_k^{\text{low}}$ between mean residual-stream activations under \emph{high} (target items at maximum, others randomized) and \emph{low} (target items at minimum) conditions, averaged over 10 random backgrounds and 20 open-ended Chinese probe questions per condition (App~\ref{app:geometry}). We compare pairwise cosine similarity between the 16 directions to the human subscale correlation matrix via the Mantel test \citep{mantel1967detection} (10{,}000 permutations).

\subsection{Psychometric geometry emerges with scale and instruction tuning}

\begin{figure}[t]
\centering
\includegraphics[width=0.40\linewidth]{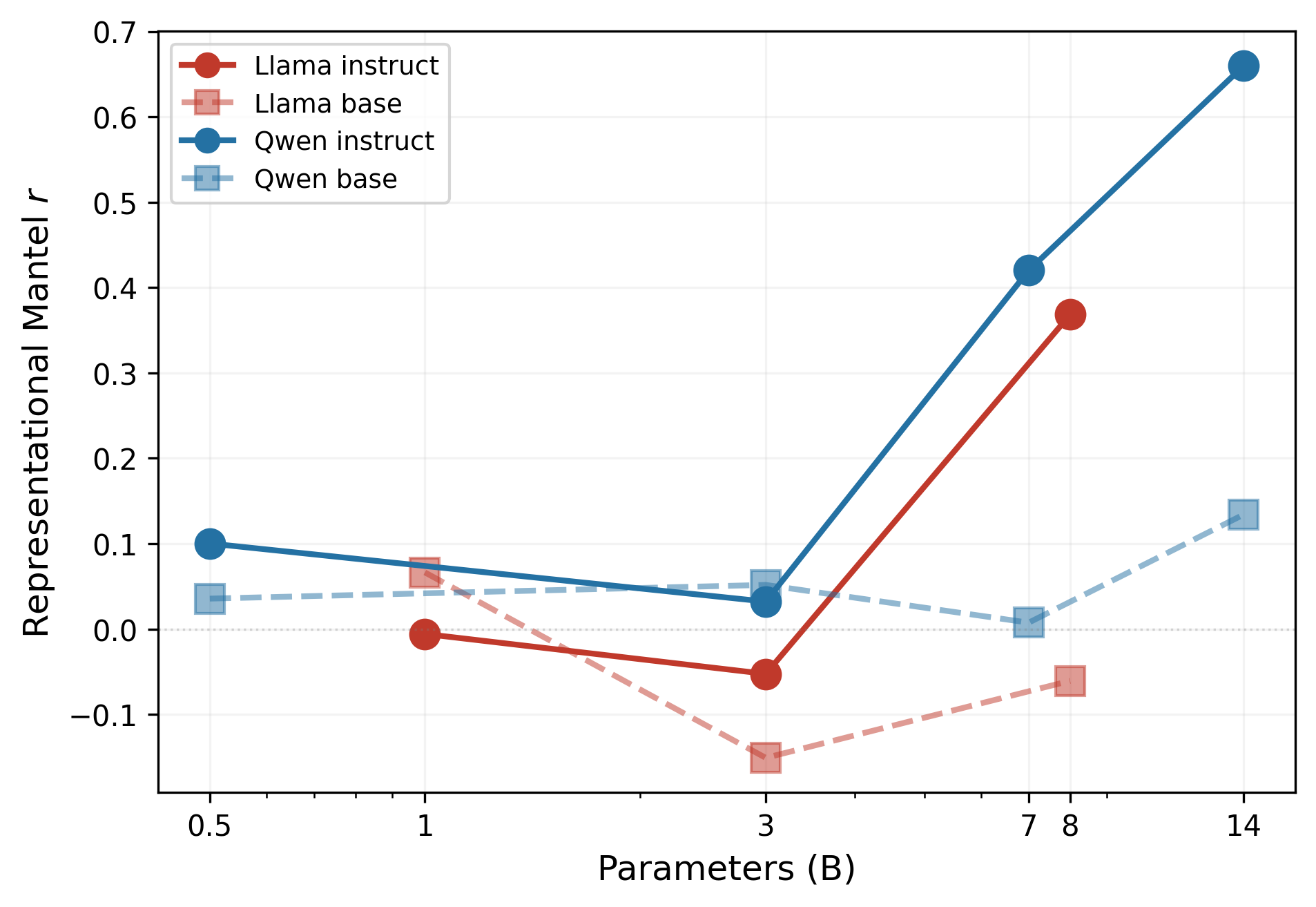}
\caption{Representational psychometric geometry across 14 variants. Mantel $r$ between contrastive-direction cosine similarity and the human subscale correlation matrix, as a function of model size. This assay detects its strongest human alignment in the tested Qwen~7B/14B and Llama~8B instruct variants; low direction reliability limits interpretation of small-model estimates.}
\label{fig:representational}
\end{figure}

Our results reveal a striking dissociation (Figure~\ref{fig:representational}). Small models (0.5B--3B) show little psychometric geometry under this assay (Mantel $r \approx 0$) regardless of instruction tuning. In contrast, large instruct models show clear structure: Llama~8B Instruct achieves Mantel $r = 0.37$, Qwen~7B Instruct $r = 0.42$, and Qwen~14B Instruct $r = 0.66$. All seven tested base variants have near-zero measured geometry (Mantel $r \leq 0.14$), including Qwen~14B Base ($r = 0.13$) and Llama~8B Base ($r = -0.06$). In these two families, size and instruction tuning are associated with the measured geometry. Across 14 variants, controlling for item-text semantic similarity reduces the mean alignment between contrastive direction geometry and human structure from $r = 0.115$ to partial $r = 0.113$ (a $2\%$ drop), with maximum absolute drop below $0.03$ across models (App~\ref{app:geometry}).

Prompt-control geometry remains aligned with the human matrix on the matched Qwen~14B Instruct subset: Mantel $r=0.481$ without scale names and $r=0.497$ with the chat template. These checks use three random-background replications rather than ten in the main analysis, so their absolute $r$ values should not be compared as equal-precision estimates. A fixed 60\%-depth layer rule also retains positive geometry for Qwen~14B Instruct (Mantel $r=0.510$); across the 14 variants the mean absolute change from the selected layer is $0.035$ (App~\ref{app:prompt_controls}).

\subsection{Convergent evidence from steering specificity}

As an exploratory intervention, we fit regression-based directions (one per subscale) from 272 subjects' activations and add scaled direction vectors to the residual stream during inference. For each (source, target) subscale pair, we record the change in target-item expected-value response. The regression-direction run yields a $16 \times 16$ steering-response matrix. In one Llama~8B Instruct run, the double-centered Mantel test (which removes generic row/column shifts to isolate construct-specific structure) against the human psychometric matrix gives $r = 0.28$, $p = 0.009$ (App~\ref{app:steering}). A separate $9 \times 16$ contrastive-direction run gives $r = 0.14$, $p = 0.216$ under the same test. Across per-model steering runs, double-centered specificity is positive in multiple instruct models, with the strongest effects observed in Llama~3B and Qwen~3B Instruct; however, magnitudes do not scale monotonically with model size and are sensitive to layer and extraction choices. These interventions show that changing fitted directions can affect responses in this steering setup; they do not demonstrate mediation of the cross-persona task.

\section{Locating Amplification in Representations}
\label{sec:amplification}

\subsection{Activation probing: paradigm and pipeline}
\label{sec:pipeline}

We now ask whether cross-scale information can be decoded from the model's activations. We design an activation probing paradigm that mirrors the behavioral experiment: the system prompt contains human responses from one scale, and we extract the decision-point activation (the last token position before the model produces its answer) when presented with a target item from a different scale. Crucially, the target scale's answers are \emph{not} in the prompt. For each target scale, we select the first 2 items per subscale to cover all 16 subscales (22--30 target items per rotation), and extract activations for 272 subjects $\times$ 7 rotations across all models.

\paragraph{Cross-scale information is linearly decodable from activations.}
Using 5-fold cross-validated ridge regression to predict each target subscale score from activations, most cross-scale subscale predictions are statistically significant ($p < 0.05$). For Llama~8B Instruct, 78 of 96 subscale-level cross-scale predictions reach significance with mean CV $r = 0.29$, supporting linear decodability of cross-scale information at the sampled layers.

\paragraph{Matched-pipeline comparison.}
To enable fair comparison between representational and behavioral amplification, we construct parallel pipelines that differ only in the item-level response extraction step:
\begin{itemize}[nosep]
\item \textbf{Behavioral pipeline}: For each target item, extract the argmax over response option log-probabilities (discrete integer). Average item responses within each subscale to obtain subscale scores. Compute the $16 \times 16$ correlation matrix of subscale scores across subjects.
\item \textbf{Representational pipeline}: For each target item, use cross-validated ridge regression to predict the item's continuous response from the decision-point activation. Average predicted item responses within each subscale (applying the same reverse coding). Compute the $16 \times 16$ correlation matrix of predicted subscale scores across subjects.
\end{itemize}
The two pipelines differ only at the item-level extraction step (continuous ridge prediction vs.\ discrete argmax); both aggregate to subscale scores via averaging, preserving the same variance structure.

\subsection{Amplification at the representational level}
\label{sec:amplification_results}

\begin{figure}[t]
\centering
\includegraphics[width=0.65\linewidth]{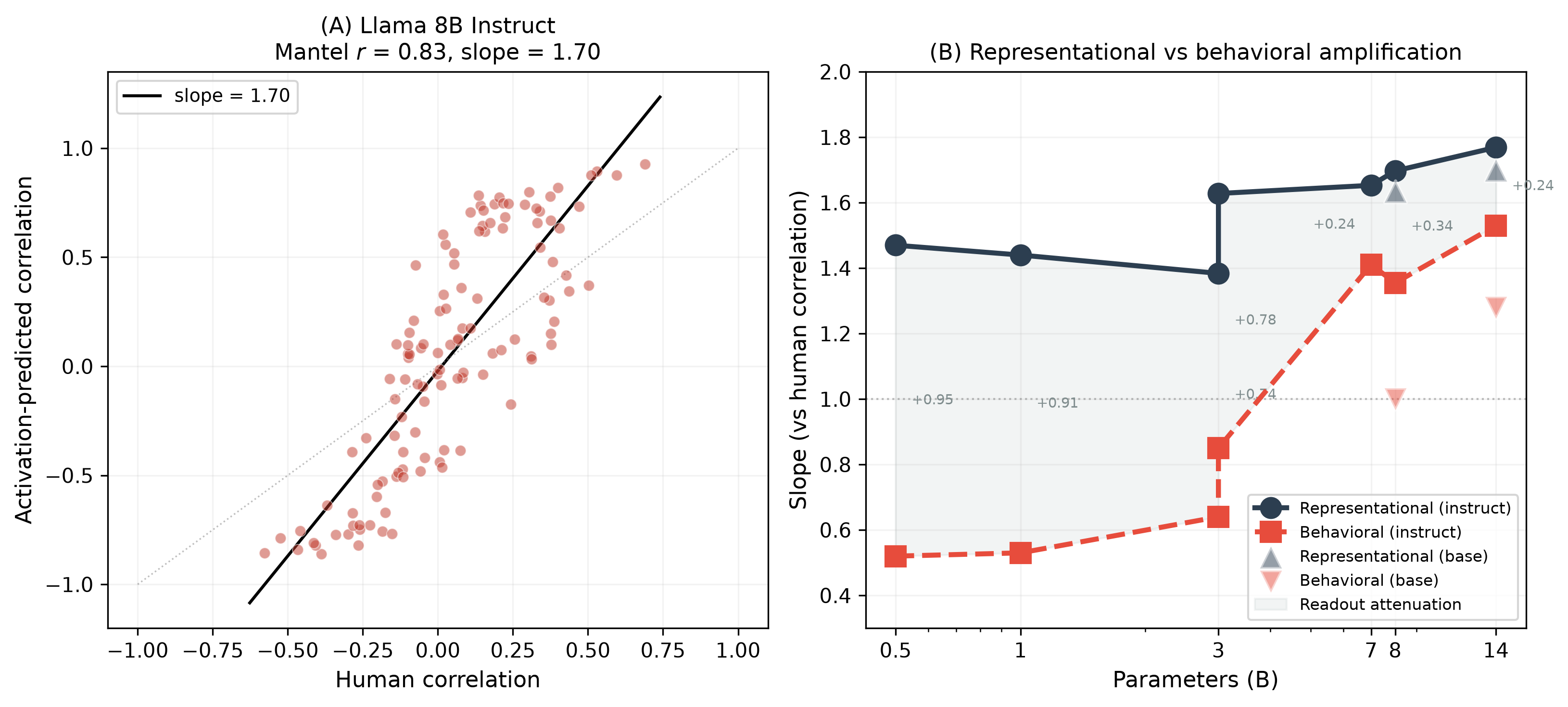}
\caption{Activation-probe and behavioral slopes across nine models. \textbf{Left}: Activation-predicted subscale correlations vs.\ human correlations for Llama~8B Instruct using the item-level pipeline (Mantel $r = 0.83$, slope $= 1.70$). \textbf{Right}: Probe-derived slopes (solid) and behavioral slopes (dashed). The probe slopes span $1.38$--$1.77$ (spread $0.39$); behavioral slopes span $0.52$--$1.53$ (spread $1.01$).}
\label{fig:amplification}
\end{figure}

Using the item-level representational pipeline (\S\ref{sec:pipeline}), we compute the predicted correlation matrix for each of nine models spanning two families (0.5B--14B), both instruction-tuned and base variants, including base--instruct pairs at 8B (Llama) and 14B (Qwen) (Table~\ref{tab:cross_persona}).

\begin{table}[t]
\centering\footnotesize
\caption{Activation-probe versus behavioral slopes across nine models. Probe slope uses the item-level pipeline (\S\ref{sec:pipeline}); behavioral slope averages seven per-scale regressions on argmax responses. Mantel $r$ compares the probe-predicted and human correlation matrices. Layers were selected post-hoc; item-level layer sweeps cover six models (App~\ref{app:robustness}).}
\label{tab:cross_persona}
\begin{tabular}{lccccc}
\toprule
Model & Selected layer & Probe slope & Beh.\ slope & Gap & Mantel $r$ \\
\midrule
Qwen 0.5B Instruct & 8  & 1.47 & 0.52 & +0.95 & 0.73 \\
Llama 1B Instruct  & 8  & 1.44 & 0.53 & +0.91 & 0.81 \\
Qwen 3B Instruct   & 14 & 1.38 & 0.64 & +0.74 & 0.81 \\
Llama 3B Instruct  & 14 & 1.63 & 0.85 & +0.78 & 0.84 \\
Qwen 7B Instruct   & 18 & 1.65 & 1.41 & +0.24 & 0.82 \\
Llama 8B Instruct  & 16 & 1.70 & 1.35 & +0.34 & 0.83 \\
Llama 8B Base      & 14 & 1.63 & 1.00 & +0.63 & 0.82 \\
Qwen 14B Instruct  & 32 & 1.77 & 1.53 & +0.24 & 0.84 \\
Qwen 14B Base      & 28 & 1.70 & 1.28 & +0.42 & 0.82 \\
\bottomrule
\end{tabular}
\end{table}

The central finding is that activation-probe slopes vary substantially less than behavioral slopes ($1.38$--$1.77$ vs $0.52$--$1.53$). The measured gap, which we term \emph{readout attenuation}, decreases across the tested instruct variants: from $+0.91$ (Llama~1B) to $+0.24$ (Qwen~14B). On Llama~8B Instruct, the item-level slope stays above $1.5$ throughout the cross-validation-selected ridge range $\lambda \in [1,100]$. In 10 item-level activation-shuffle runs, the slope falls from $1.70$ to $0.25 \pm 0.02$ SD. A separate subscale-level control permutes subject score labels 100 times. None exceeds its observed slope of $1.96$; the permutation mean is $0.70 \pm 0.20$ SD ($p < 0.01$; App~\ref{app:robustness}). These controls support subject-specific cross-scale information in the activations. The supervised probe and discrete output remain different measurements.

The two base models are informative: activation-probe slopes are comparable to their instruct counterparts (Llama~8B: $1.63$ vs.\ $1.70$; Qwen~14B: $1.70$ vs.\ $1.77$), but behavioral slopes are lower ($1.00$ vs.\ $1.35$; $1.28$ vs.\ $1.53$). Extending to all seven base--instruct pairs, instruction tuning is associated with higher behavioral slopes (mean $\Delta=+0.22$, paired $p = 0.011$). The geometry gap is larger for the tested pairs $\geq 7$B (mean $+0.46$) than for smaller pairs (mean $+0.02$). These measurements do not identify the operations behind the probe--output gap or a size threshold.

\paragraph{Discretization and readout decomposition.}
\label{sec:discretization}
Comparing continuous ridge, rounded-integer, and argmax pipelines (mean slopes $1.60$, $0.71$, $1.01$), behavioral slope typically \emph{exceeds} the rounded slope. The difference is near zero in small variants but $+0.56$ in Qwen~14B Instruct (rounded $0.97 \to$ behavioral $1.53$; App~\ref{app:robustness}).

\subsection{Synthetic pipeline control}
\label{sec:pipeline_control}

Is the $1.38$--$1.77$ probe-slope range a property of learned representations, or partly an artifact of the ridge pipeline? We address this question using a synthetic control with known ground-truth correlation structure. Specifically, we generate $N = 272$ latent subscale scores from the human $16\times 16$ correlation matrix, project them through random Gaussian matrices $\mathbf{W} \in \mathbb{R}^{d \times 16}$, and add isotropic noise at SNR levels $\in \{0.1, 0.3, 1.0, 3.0, 10.0\}$ across 20 random seeds. We then apply the same item-level ridge pipeline (App~\ref{app:pipeline_control}).

\begin{figure}[t]
\centering
\includegraphics[width=0.60\linewidth]{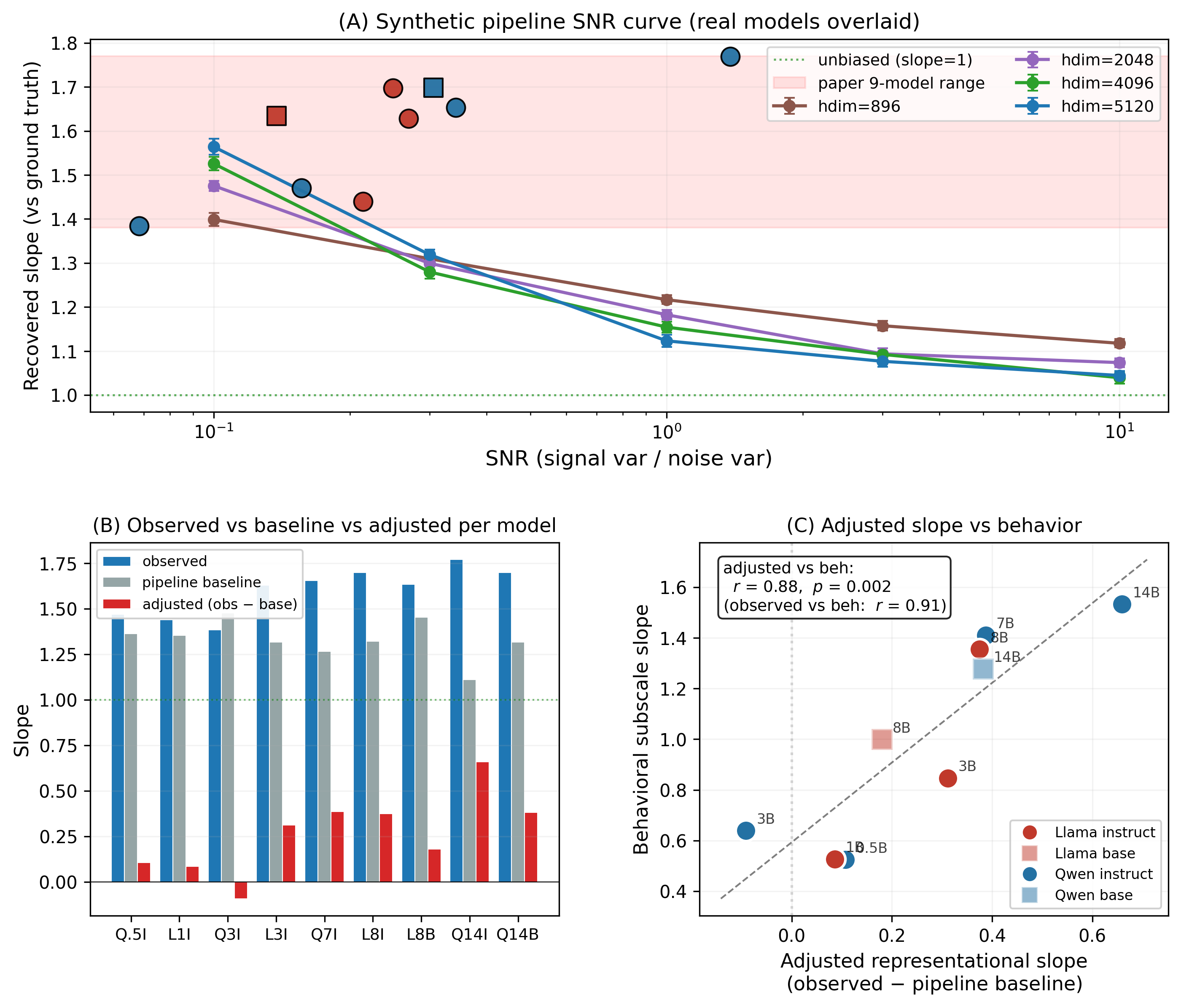}
\caption{Synthetic pipeline control. \textbf{(A)}: Recovered slope of the item-level ridge pipeline on synthetic data as a function of SNR, for four hidden dimensions. Slope approaches $1.0$ at high SNR and rises to $\approx 1.5$ at low SNR. Real models are overlaid using an SNR proxy from contrastive-direction reliability, a different pipeline; most fall within the nine-model observed slope range ($1.38$--$1.77$, red band). \textbf{(B)}: Per-model observed slope (blue), proxy-based synthetic baseline (grey), adjusted slope (red). \textbf{(C)}: Adjusted slope remains associated with behavioral slope ($r = 0.88$, $p = 0.002$; raw $r = 0.91$).}
\label{fig:pipeline_control}
\end{figure}

Pipeline inflation is SNR-dependent, not a fixed offset: recovered slope is $\approx 1.05$ at SNR $= 10$ but $\approx 1.5$ at SNR $= 0.1$ (Fig.~\ref{fig:pipeline_control}A). Mantel $r$ between recovered and ground-truth matrices remains $> 0.97$ throughout, so the pipeline preserves correlation \emph{structure} faithfully while inflating magnitude. To estimate a baseline for each model, we interpolate the synthetic grid using contrastive-direction reliability $\rho$ as an SNR proxy ($\text{SNR}\approx\rho/(1-\rho)$). The resulting baselines span $1.11$--$1.48$. This reliability measure comes from a different pipeline, so the baselines are not matched corrections for the item-level probe. Proxy-adjusted slopes (observed $-$ baseline) span $-0.09$ to $+0.66$. Across nine variants, they correlate with behavioral slope at $r=0.88$ ($p=0.002$), compared with $r=0.91$ for raw slopes.

\section{From Representations to Behavior}
\label{sec:linking}

\begin{figure}[t]
\centering
\includegraphics[width=0.40\linewidth]{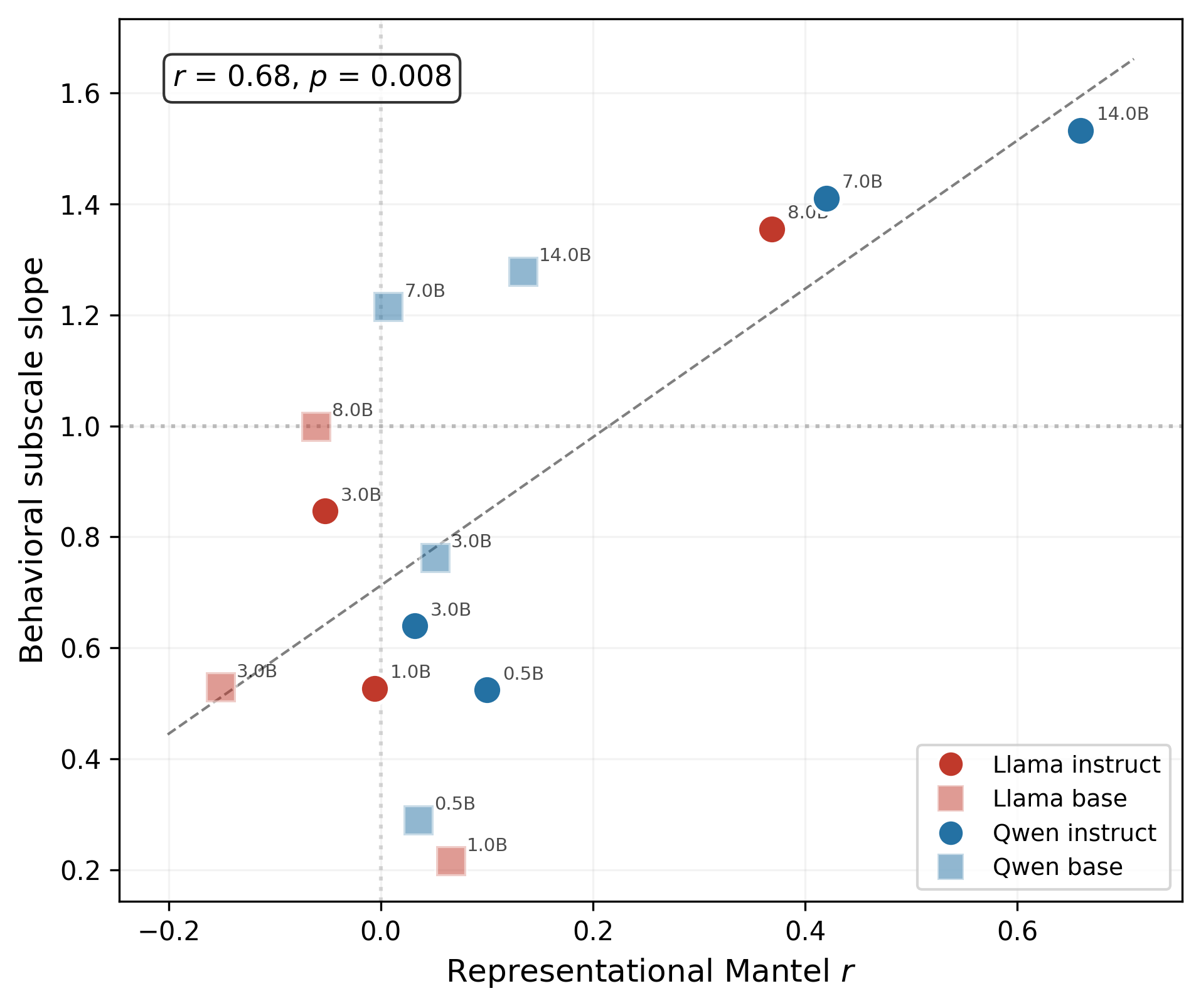}
\caption{Representational geometry is associated with behavioral amplification across 14 variants ($r = 0.68$, $p = 0.008$; leave-one-model-out: [0.58, 0.74]). Circles = instruct, squares = base; red = Llama, blue = Qwen; labels indicate model size.}
\label{fig:structure_behavior}
\end{figure}

If amplification is rooted in representations, models with stronger representational structure should exhibit greater behavioral amplification. Across all 14 variants, representational Mantel $r$ correlates with behavioral subscale slope at $r = 0.68$ ($p = 0.008$; Fig.~\ref{fig:structure_behavior}). A model-resampling 95\% CI is $[0.19,0.87]$ and the leave-one-model-out range is $[0.58,0.74]$. Controlling for log model size gives partial $r = 0.69$ (95\% CI $[0.34,0.91]$; leave-one-model-out range $[0.63,0.79]$). Variants within a family are related, so these intervals do not establish generalization to unrelated families. Within instruct variants, $r = 0.90$ ($p = 0.006$); within base variants, $r = 0.17$ ($p = 0.71$). Each group has seven variants; the Fisher-$z$ comparison gives $p = 0.064$.

\paragraph{A base--instruct dissociation.}
Qwen~14B Base has a behavioral slope of $1.28$ despite little directional geometry by this assay (Mantel $r = 0.13$). Llama~8B Base has a slope near $1.00$ and similarly low measured geometry ($r = -0.06$). Behavioral amplification can therefore coexist with low alignment in this geometry assay; whether base and instruct variants use different causal pathways remains unresolved.

\section{Discussion}
\label{sec:discussion}

\paragraph{Interpreting probe slopes.}
Our central finding is that activation-probe slopes vary less than behavioral slopes across models ($1.38$--$1.77$ vs $0.52$--$1.53$). The measured gap is smaller in larger instruct variants. A synthetic pipeline control (\S\ref{sec:pipeline_control}) shows that ridge probing can inflate slopes. After subtracting a proxy-based baseline, adjusted slopes span $-0.09$ to $+0.66$ and remain associated with behavioral slope ($r = 0.88$). The proxy comes from contrastive-direction reliability, not the item-level probe. Thus the adjustment does not precisely estimate intrinsic representational amplification.

\paragraph{Relation to prior interpretations.}
\citet{liu2026five} interpret structural amplification as a form of noise filtering, where unilateral attenuation correction should bring slopes toward $1.0$. Our results complicate this account. Corrected behavioral slopes remain above $1.0$ for models $\geq 7$B ($1.19$--$1.39$), whereas Liu et al.\ report clustering near $1.0$. One potential source of divergence is methodological: our rotating-persona design spans the full bidirectional correlation structure across constructs, whereas their single-source design projects from Big Five inputs onto other scales. The residual amplification we observe may therefore arise from correlations not directly constrained by a Big Five input. In addition, activation-probe slopes exceed $1.0$ even in smaller models whose behavioral outputs are attenuated below $1.0$; the two slopes are measured by different pipelines.

\paragraph{Emergence, instruction tuning, and implications.}
Instruction tuning is associated with two patterns. First, organized directional geometry is observed in the larger tested instruct variants (Qwen~14B base vs.\ instruct: Mantel $r=0.13$ vs.\ $0.66$), but not in the tested base variants under this assay. Second, base and instruct variants exhibit similar activation-probe slopes but differ in behavioral expression. Extending this comparison across seven base--instruct pairs clarifies the pattern: instruct variants have higher behavioral slopes (paired $p=0.011$), whereas improvements in representational geometry are observed mainly above $\approx 7$B. Smaller instruct models therefore show behavioral gains without comparable geometric reorganization under this assay. This pattern aligns with the dissociation observed in base models (\S\ref{sec:linking}). Base models appear to encode linearly decodable psychometric information without organized directional geometry under this assay. More broadly, psychometric structure provides a quantitative and population-grounded testbed for examining how structured representations emerge in LLMs.

\paragraph{Why instruction tuning aligns with personality structure.}
These findings raise a further question: why does instruction tuning align representations with \emph{human} personality structure, despite not explicitly targeting such constructs? One possibility is incidental trait alignment. Signals related to human preference and role consistency, which are abundant in instruction-tuning data, may organize latent personality-relevant representations already present in the base model into directions that happen to align with human trait structure. Under this view, structured trait representations may emerge without explicit supervision, as a byproduct of optimization toward other objectives. This possibility remains untested here.

\paragraph{Limitations.}
Our human data are drawn from a single cultural sample (Taiwanese college students assessed in Traditional Chinese). Generalization to other populations or languages requires matched human samples; translating model prompts alone would not provide that comparison. An exploratory post-hoc analysis (App~\ref{app:cultural}) suggests that culturally specific subscales exhibit stronger amplification, although this effect is driven largely by a small number of subscale pairs and requires replication. Importantly, amplification is not confined to culturally specific constructs: within culturally stable subscales alone, $\geq 3$B instruct models still exhibit slopes above $1.0$. This analysis is not a cross-cultural replication. The 14 variants also come from two model families, and variants within a family are not independent training replications. Contrastive direction reliability is limited in smaller models, constraining interpretation of near-zero geometry results. Steering was exploratory and did not test mediation in the cross-persona task. Finally, the SNR proxy is imperfect, and adjusted slope magnitudes should be interpreted as approximate.

\paragraph{Broader impacts.}
These measurements may help evaluate how LLMs simulate psychological response patterns. Psychometric profiling could also be misused or applied to individuals without consent; this study evaluates predictions conditioned on participants' supplied questionnaire responses and does not validate individual-level assessment in deployment.

\paragraph{Code and data availability.}
The \href{https://github.com/CHEN-lab-NTU/Tracing-Psychometric-Inference-in-Large-Language-Models}{project repository} contains the experiment code, anonymized participant-level item responses, and the Traditional Chinese questionnaires administered in this study.

\section{Conclusion}
\label{sec:conclusion}

We trace psychometric inference in LLMs from behavioral outputs to internal representations. Structured psychometric geometry is detected most clearly in larger instruct variants and is associated with behavioral amplification across models ($r=0.68$; partial $r=0.69$ controlling for log size). Activation-probe slopes span $1.38$--$1.77$, whereas behavioral slopes span $0.52$--$1.53$. A synthetic control shows that this probe range includes ridge-pipeline inflation; after proxy-based subtraction, adjusted slopes remain associated with behavior ($r=0.88$). The results distinguish decodability, geometry, and output behavior. They do not identify a causal path from the measured directions to cross-persona answers.

\section*{Acknowledgements and Disclosures}
\textbf{Author contributions.} Chih-Hao Hsu led the project, proposed the representation-level research question, designed the mechanistic experiments and robustness controls, developed the code, conducted the experiments and analyses, interpreted the results, wrote the manuscript, and handled the rebuttal. Feng-Chun B. Chou proposed the behavioral research questions and the cross-persona paradigm. Pin-Hao A. Chen proposed the behavioral research question and provided the human survey data as well as computing resources. All authors reviewed the manuscript.\\

\textbf{Acknowledgements.} Feng-Chun B. Chou was supported by the Fellowship for Doctoral Students in Humanities and Social Sciences from Academia Sinica, Taiwan.

\textbf{Funding.} This research was supported by funding from the National Science and Technology Council and National Taiwan University in Taiwan (NSTC 111-2423-H-002-008-MY4, NSTC 115-2628-H-002-029-MY3, NTU-114L7870 and NTU-115L7846 to P.-H.A. Chen).

\textbf{Competing interests.} The authors declare no conflicts of interest.

\bibliographystyle{plainnat}
\bibliography{references}

\appendix

\section{Prompt Examples}
\label{app:prompts}

\paragraph{Single-scale system prompt (behavioral experiment and cross-persona probing).}
\begin{verbatim}
You are completing psychological questionnaires as a person
with the following responses on the [Scale Name]:
[instruction]
Response scale:
  1 = [label] ... K = [label]

Q1. [item text] -> [response]
Q2. [item text] -> [response]
...

Respond to all questions as this person would, maintaining
consistency with this psychological profile. Answer with just
the number corresponding to your choice.
\end{verbatim}

\paragraph{Contrastive extraction system prompt (mega-prompt).}
Same format but includes all 7 scales (114 items). The target subscale's items are set to maximum (high) or minimum (low); all other items are randomized.

\paragraph{Target item prompt.}
\begin{verbatim}
[instruction]

Statement: [item text]

Response options:
  1 = [label] ... K = [label]

Please respond with a single number from 1 to K.
\end{verbatim}

Here $K$ and the response labels follow the administered instrument (Table~\ref{tab:scales}). The bracketed fields vary by instrument or item; the administered questionnaire wording is included in the accompanying repository.

\section{Prompt, Readout, and Fixed-Layer Controls}
\label{app:prompt_controls}

The prompt and readout controls use a matched 50-participant subset of Qwen~14B Instruct. The table reports between-scale subscale pairs; valid-pair counts differ across readouts. These controls cover one model rather than a complete 14-variant robustness sweep.

Under the standard cross-persona prompt, expected-value readout gave between-scale subscale alignment $r=0.727$ and slope $1.438$ on the same subset. The table below reports the two alternative prompt formats and their valid-pair counts.

\begin{center}\small
\begin{tabular}{llccc}
\toprule
Prompt & Readout & Alignment $r$ & Slope & Valid pairs \\
\midrule
No scale name & Argmax & 0.753 & 1.401 & 98 \\
No scale name & Expected value & 0.752 & 1.497 & 102 \\
Chat template & Argmax & 0.766 & 1.330 & 94 \\
Chat template & Expected value & 0.736 & 1.338 & 102 \\
\bottomrule
\end{tabular}
\end{center}

Using the same prompt variants, the contrastive-geometry controls yield Mantel $r=0.481$ ($p=0.0002$) without scale names and $r=0.497$ ($p=0.0001$) with the official chat template. Both use three random-background replications instead of the ten used in the main analysis. In a separate factorial Big Five control with 243 synthetically specified profiles, Qwen~14B Instruct's profile geometry has Mantel $r=0.476$ ($p=0.001$), and manipulated factors are linearly readable from activations (mean $r=0.996$). This establishes sensitivity to controlled persona variation, not the absence of learned psychological priors.

We also repeat the geometry analysis at the sampled layer nearest 60\% depth in each model. Across 14 variants, the mean absolute change from the selected layer is $0.035$ for Mantel $r$ and $0.031$ for geometry slope. Qwen~14B Instruct changes from $r=0.660$ to $0.510$, and Qwen~7B Instruct from $0.420$ to $0.294$. Llama~8B Instruct stays at $r=0.368$: its selected layer~18 is also the nearest sampled layer to 60\% depth. Layer choice matters for some models.

\section{Robustness Analyses}
\label{app:robustness}

\begin{table}[H]
\centering\footnotesize
\setlength{\tabcolsep}{3pt}
\caption{Uncertainty for key results. Behavioral intervals use 5{,}000 paired-participant bootstrap samples; geometry intervals resample matrix pairs; across-model intervals resample variants. The leave-one-out (LOO) unit is participant, construct, or variant, respectively. The Qwen geometry rows use layer~36. Matrix pairs and model variants are dependent, so their intervals are descriptive.}
\label{tab:uncertainty}
\begin{tabular}{p{0.39\linewidth}ccc}
\toprule
Statistic & Estimate & 95\% interval & LOO range \\
\midrule
Qwen~14B Instruct: behavioral slope & 1.533 & [1.323, 1.659] & [1.513, 1.565] \\
Qwen~14B Instruct: behavioral alignment $r$ & 0.809 & [0.748, 0.833] & [0.802, 0.814] \\
Qwen~14B Instruct: geometry Mantel $r$ & 0.660 & [0.567, 0.747] & [0.626, 0.706] \\
Qwen~14B Instruct: geometry slope & 0.765 & [0.601, 0.939] & [0.687, 0.902] \\
14 variants: geometry--behavior $r$ & 0.678 & [0.186, 0.872] & [0.576, 0.737] \\
14 variants: partial $r$ (log size) & 0.691 & [0.343, 0.911] & [0.627, 0.788] \\
\bottomrule
\end{tabular}
\end{table}

\paragraph{Bootstrap stability and split-half baseline for alignment metrics.}
Bootstrap resampling (2000 iterations, subject-level) of the human $16 \times 16$ subscale correlation matrix yields mean pairwise CI width of $0.24$ (median $0.24$, max $0.31$). The overall structure is highly stable ($r = 0.97$ between original and bootstrap samples; 95\% CI: $[0.96, 0.99]$).

\begin{figure}[t]
\centering
\includegraphics[width=\linewidth]{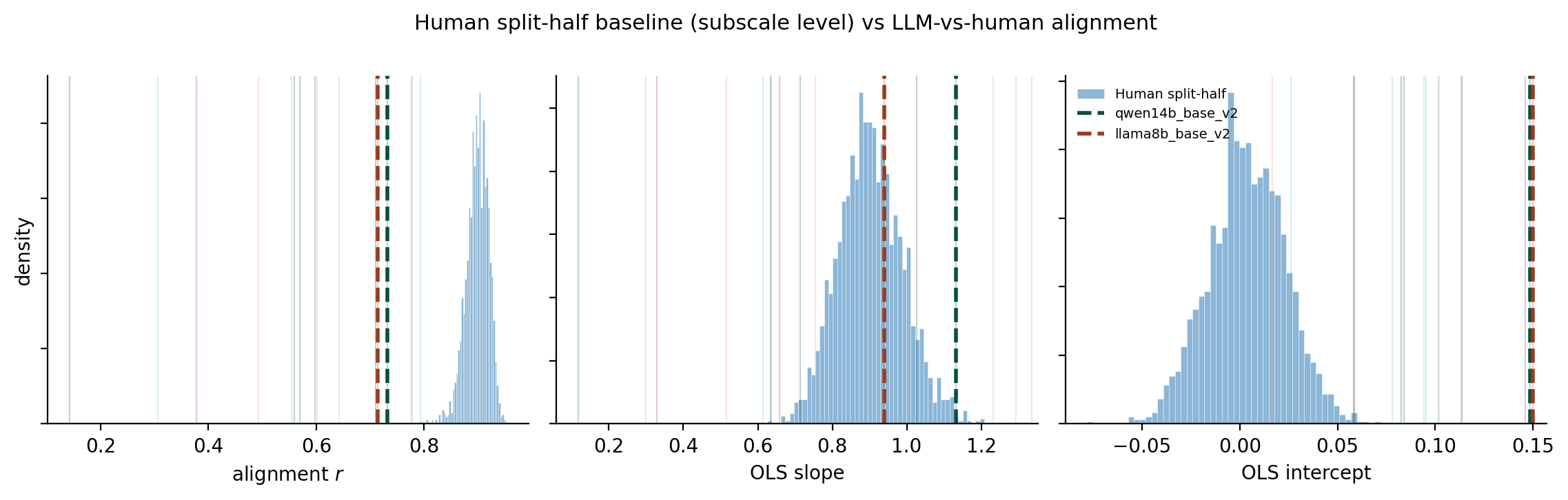}
\caption{Human split-half baseline at the subscale level ($2000$ random splits of $N = 272$, blue histograms) versus two pooled LLM base-model matrices (Qwen~14B Base, green; Llama~8B Base, red), for alignment $r$ (left), OLS slope (middle), and OLS intercept (right). Both LLMs fall \emph{below} the human distribution on $r$ and \emph{above} it on intercept. On slope, Qwen~14B Base is near the upper edge, while Llama~8B Base lies within the distribution. The three panels correspond to three independent components of the LLM-vs-human relationship, and the LLM offsets are signed differently across them.}
\label{fig:split_half_baseline}
\end{figure}

To establish a reference scale for our LLM-vs-human alignment metrics for the pooled 16-subscale comparison, we additionally compute a split-half human baseline: 2000 random partitions of the 272 subjects into two halves, each yielding an independent $16 \times 16$ subscale correlation matrix, with the alignment statistics ($r$, OLS slope, OLS intercept) computed between the two halves. This distribution characterises how closely two independent human samples reproduce each other under the same metric, and provides a natural reference against which to compare LLM-vs-human alignment (Figure~\ref{fig:split_half_baseline}). Three observations follow. (i) On alignment $r$, our 8B/14B base LLM matrices fall in the left tail of the human split-half distribution. LLM cross-scale predictions therefore approach but do not yet reach the level of agreement that two independent halves of the human sample show with one another, framing the LLM alignment numbers reported in the main text as substantial but sub-ceiling. (ii) On OLS slope, Qwen~14B Base lies near the upper edge of the human distribution, whereas Llama~8B Base lies within it. The human reference is centred near $1.0$ by construction. (iii) On intercept, both LLM models lie beyond the right tail of the human distribution. We discuss this intercept offset, which is independent of the slope amplification, in App~\ref{app:behavioral} (Intercept inflation).

\paragraph{Leave-one-out analysis on structure--behavior correlation.}
The $r = 0.68$ correlation between representational Mantel $r$ and behavioral slope (Figure~\ref{fig:structure_behavior}) is robust to removal of individual models. LOO range: [0.58, 0.74]; LOO mean: 0.68. The most influential point is Qwen~14B Instruct (removal reduces $r$ to 0.58).

\paragraph{Pipeline sensitivity analysis (Llama~8B Instruct).}
Three representational pipeline variants are compared:
\begin{center}\small
\begin{tabular}{lcc}
\toprule
Pipeline & Slope & Mantel $r$ \\
\midrule
Ridge $\to$ subscale score (continuous) & 1.96 & 0.92 \\
Ridge $\to$ item (continuous) $\to$ subscale mean \textbf{[main]} & 1.70 & 0.83 \\
Ridge $\to$ item $\to$ round to integer $\to$ subscale mean & 0.83 & 0.78 \\
Behavioral (argmax integer $\to$ subscale mean) & 1.35 & --- \\
\bottomrule
\end{tabular}
\end{center}
The item-level continuous pipeline (second row) is used as the main representational pipeline throughout the paper, as it provides an apples-to-apples comparison with the behavioral pipeline (both aggregate item-level responses to subscale scores via averaging).

\paragraph{Discretization across all 9 models.}
The continuous $\to$ rounded-integer slope reduction reported above on Llama~8B Instruct ($1.70 \to 0.83$) is consistent across the full 9-model panel. Continuous slopes range $1.38$--$1.77$ (mean $1.60$); rounded slopes range $0.47$--$0.97$ (mean $0.71$); within-model differences range $+0.80$ to $+0.97$. Paired $t$-test on the 9-model paired differences yields $p < 10^{-4}$ (Wilcoxon $p = 0.004$). Mantel $r$, being scale-invariant, is largely preserved after discretization and in two models (Qwen~0.5B Instruct, Llama~1B Instruct) marginally increases, consistent with continuous predictions having inflated tail variance at small human correlations.

\begin{figure}[t]
\centering
\includegraphics[width=0.55\linewidth]{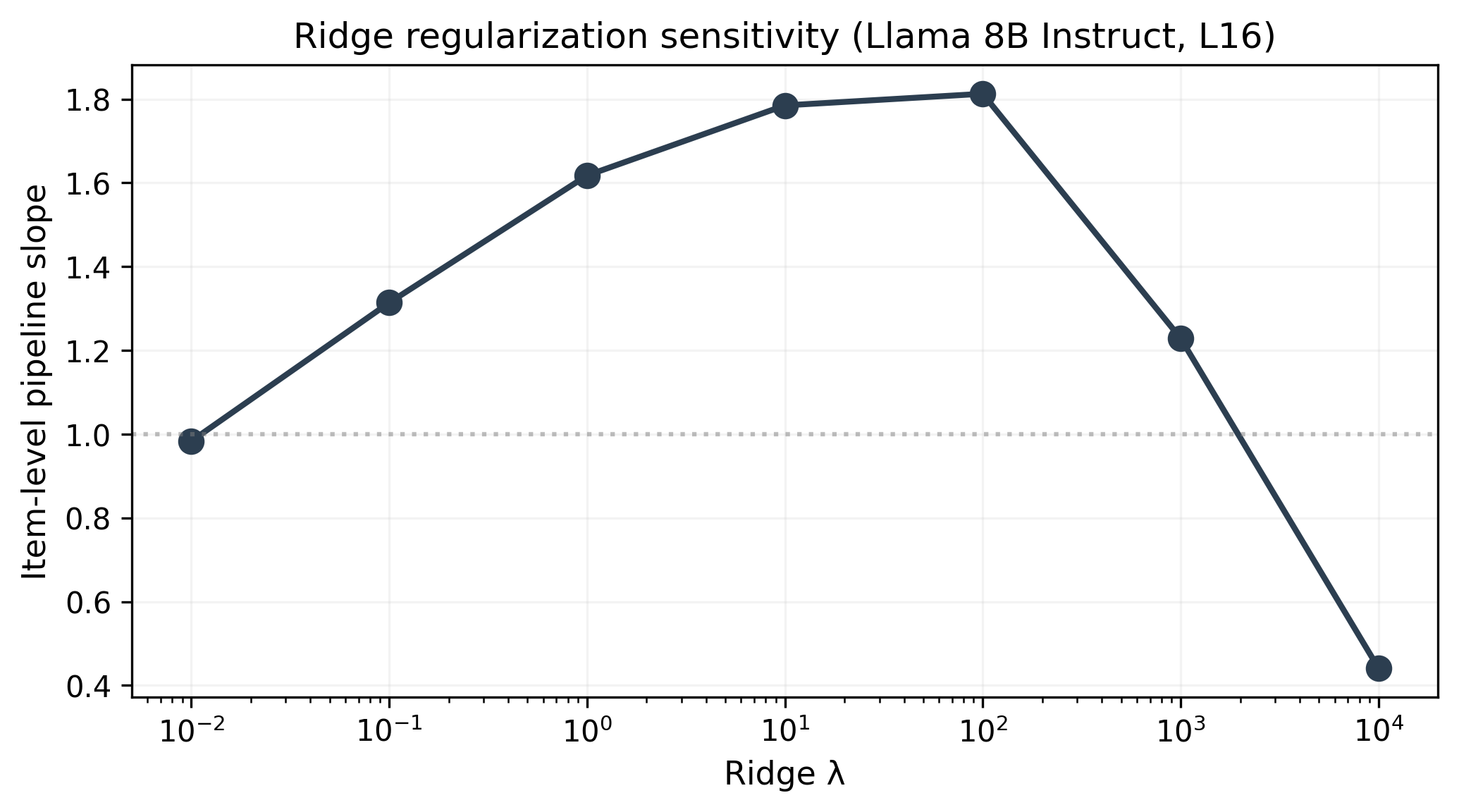}
\caption{Ridge regularization sensitivity on Llama~8B Instruct (layer~16). Item-level pipeline slope as a function of ridge $\lambda$ (log scale), exhibiting the expected inverted-U: overfitting at $\lambda \to 0$ (OLS-like, activation noise overfit) and over-regularization at $\lambda \to \infty$ (weights shrunk to zero). The CV-selected range ($\lambda \in [1, 100]$) falls on the plateau, where slope remains above $1.5$. The main result is therefore robust to the specific regularization choice within the range that cross-validation actually selects.}
\label{fig:lambda_sensitivity}
\end{figure}

\paragraph{Ridge regularization sensitivity.}
On Llama~8B Instruct (layer~16), we vary the ridge regularization parameter $\lambda$ across six orders of magnitude. The item-level pipeline slope remains above 1.5 for $\lambda \in [1, 100]$ (the range selected by cross-validation), peaking at $1.81$ ($\lambda = 100$) and declining to $0.98$ at $\lambda = 0.01$ (under-regularized) and $0.44$ at $\lambda = 10{,}000$ (over-regularized). The main result is not sensitive to the specific regularization strength within the cross-validated range (Figure~\ref{fig:lambda_sensitivity}).

\paragraph{Random features baseline.}
We shuffle the subject--activation mapping within each rotation (breaking the correspondence between persona identity and activation while preserving the marginal distribution of activations) and recompute the item-level pipeline. Over 10 shuffled iterations on Llama~8B Instruct, the mean slope is $0.25 \pm 0.02$ SD, compared to $1.70$ for real activations, a $6.7\times$ difference. This shows that the representational slope depends on subject-specific activation structure; it does not establish how the model uses that information in its output.

\begin{figure}[t]
\centering
\includegraphics[width=0.8\linewidth]{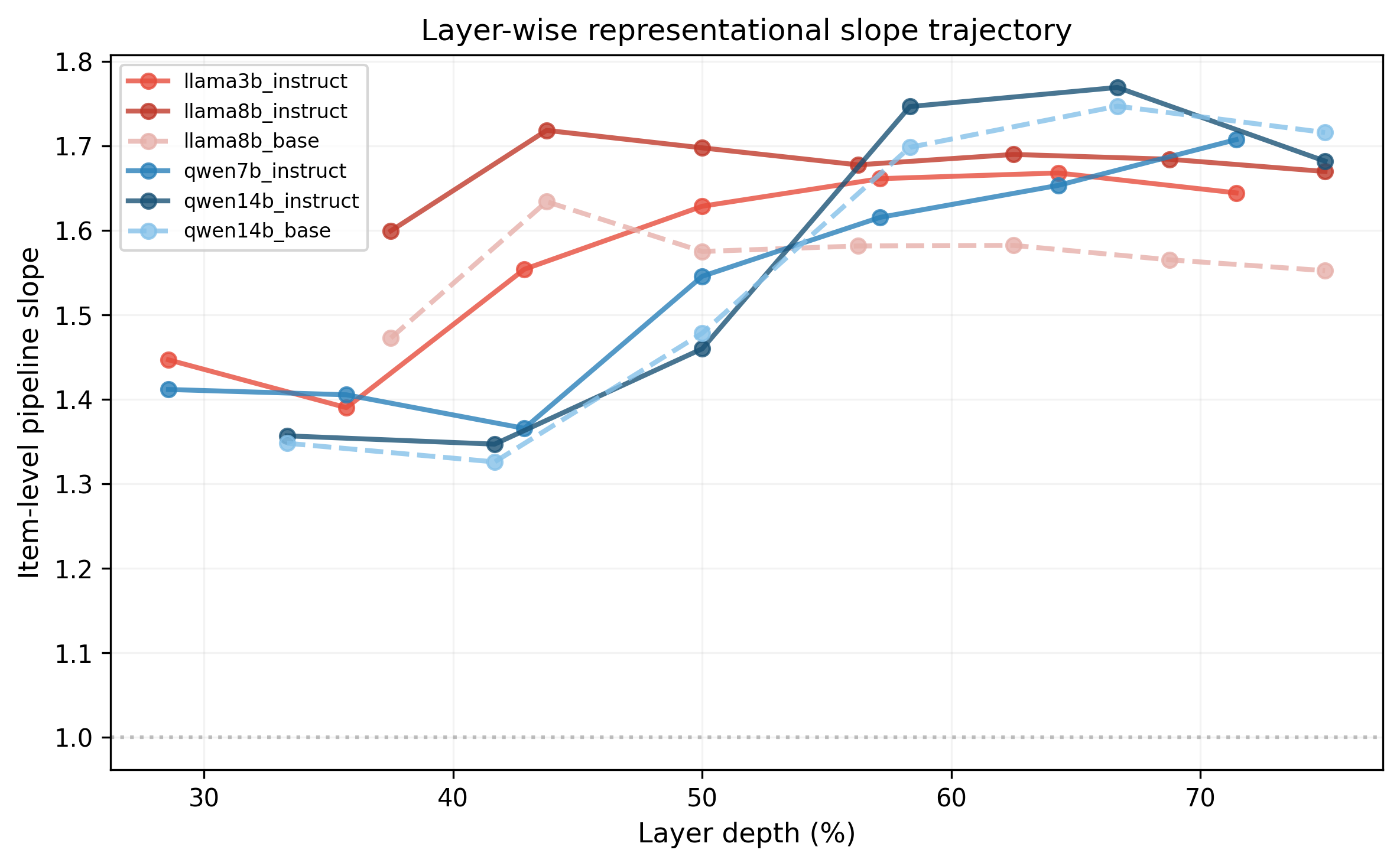}
\caption{Item-level pipeline slope by relative layer depth for the six models with layer sweeps. Circles show the tested layers, including the post-hoc selected layers in Table~\ref{tab:cross_persona}; these selected layers are not always the item-level slope maxima.}
\label{fig:layerwise}
\end{figure}

\paragraph{Layer selection and robustness.}
The layers in Table~\ref{tab:cross_persona} were selected post-hoc. A separate 14-variant check of \emph{contrastive geometry} found mean slope spreads of $0.025$ across neighboring sampled layers and $0.043$ across two sampled steps (maximum $0.118$ and $0.137$). These values do not measure the item-level probe slope; item-level sweeps cover six models (Figure~\ref{fig:layerwise}).

\paragraph{Layer-wise slope trajectory.}
The item-level pipeline slope varies across layers for the six models with layer sweeps, generally increasing from the first sampled layer ($1.35$--$1.60$) to a peak within the sampled range ($1.63$--$1.77$); five of the six decline slightly after that peak (Figure~\ref{fig:layerwise}). The slope remains elevated ($> 1.5$) through the final extracted layer of each sweep. The selected layer is marked in the plot but is not always the item-level slope maximum. These observations describe the sampled mid-to-late layers; they do not identify the model operation that creates the gap between the probe and behavioral pipelines.

\paragraph{Within-group and partial correlations.}
The $r = 0.68$ structure--behavior correlation (\S\ref{sec:linking}) was decomposed to test scale and instruct-status confounds. Within the 7 instruct models, the correlation is $r = 0.90$ ($p = 0.006$); within the 7 base models, $r = 0.17$ ($p = 0.71$); Fisher-$z$ comparison $p = 0.064$. The partial correlation controlling for log model size is $r_{\text{partial}} = 0.69$ (versus zero-order $r = 0.68$). Additional control for instruct/base status yielded $r_{\text{partial}} = 0.55$, retaining a substantial effect size. These descriptive adjustments do not resolve dependence among variants from the same families.

\section{Trivial-Context Control}
\label{app:trivial_context}

In the cross-persona paradigm, a subject's item-level responses on the persona scale are placed in the prompt. For target subscales from the \emph{same} scale as the persona (``within-scale'' subscale pairs), direct supporting context is therefore available to the model; for subscales from a different scale (``cross-scale'' pairs), no direct context exists. A natural concern is that alignment on within-scale pairs could reflect trivial prompt copying rather than structured inference, inflating the overall alignment metrics.

To test this, we partition the cross-persona predicted correlation matrices of all 9 cross-persona models into within-scale and cross-scale pairs, and compute the mean absolute deviation $|r_\text{pred} - r_\text{human}|$ for each group. Pooling across models, within-scale pairs yield mean $|r_\text{pred} - r_\text{human}| = 0.293$ (162 pairs) vs.\ cross-scale $= 0.284$ (918 pairs); the ratio is $1.030$, indistinguishable from no effect. Paired per-model testing yields $t$-test $p = 0.24$, Wilcoxon $p = 0.25$. We conclude that within-scale prompt context does not systematically improve alignment over cross-scale pairs, ruling out trivial copying of in-prompt responses as a driver of the observed structure.

\section{Cultural Specificity Exploratory Analysis}
\label{app:cultural}

As an exploratory check on whether the observed amplification is uniform across culturally stable versus culturally specific constructs, we classify subscales based on prior cross-cultural literature: \emph{stable} = Big Five E/A/C/N, IRI EC/PT, PANAS NA, SWLS (8 subscales); \emph{specific} = POM, SCS Indep/Interdep (3 subscales); remaining subscales (IRI Fantasy and Personal Distress, PANAS PA, BFI Openness, UCLA Loneliness) are marked ambiguous and excluded from the contrast because cross-cultural evidence on their replicability in East Asian samples is mixed. For each of 9 models with cross-persona data, we compute OLS slope and Mantel $r$ on the subset of subscale pairs within each group.

Paired comparisons indicate that culturally specific subscales exhibit \emph{greater} slope amplification than stable subscales (mean $\Delta$slope $= -0.42$, paired $t$-test $p < 10^{-4}$, Wilcoxon $p = 0.004$; the Mantel $r$ difference is similarly signed, $\Delta = -0.25$, paired $t$-test $p < 10^{-5}$). The direction survives restricting to cross-scale pairs only ($\Delta = -0.73$, $p < 10^{-4}$) and using the magnitude-invariant Mantel $r$ instead of slope.

However, leave-one-pair-out analysis on the three within-specific pairs reveals that this contrast is driven substantially by a single pair: POM $\times$ SCS Independent self (mean $|\Delta\text{slope}|$ across 9 models $= 1.71$). The other two within-specific pairs are SCS Indep $\times$ SCS Interdep (mean $|\Delta\text{slope}|$ $= 0.45$) and POM $\times$ SCS Interdependent self (mean $|\Delta\text{slope}|$ $= 0.12$). The small subscale count in the specific group (3 subscales $\to$ 3 within-group pairs) limits statistical generalization.

This post-hoc contrast is driven substantially by one construct pair and cannot identify the cause of the difference. It does not test whether training-language exposure, translation, or learned cultural stereotypes produced the observed pattern. Cross-cultural replication with matched human samples and expanded construct sets would be needed to evaluate generality.

\section{Synthetic Pipeline Control Details}
\label{app:pipeline_control}

\paragraph{Item-level synthetic pipeline.}
For items-per-subscale $\in \{3, 5, 7\}$ and target Cronbach's $\alpha \in \{0.60, 0.75, 0.85\}$, the $20$-seed recovered slope (vs.\ sample latent correlation) ranges from $1.17$ to $1.43$ (Table below). The mean slope across all $9 \times 20 = 180$ runs is $1.289$ (std $0.096$). Pipeline inflation decreases as items-per-subscale and $\alpha$ increase, consistent with higher measurement precision yielding lower ridge regularization and thus less information sharing across items.

\begin{center}\small
\begin{tabular}{ccccc}
\toprule
items/subscale & $\alpha = 0.60$ & $\alpha = 0.75$ & $\alpha = 0.85$ \\
\midrule
3 & $1.430 \pm 0.007$ & $1.394 \pm 0.005$ & $1.196 \pm 0.006$ \\
5 & $1.343 \pm 0.006$ & $1.259 \pm 0.003$ & $1.231 \pm 0.002$ \\
7 & $1.380 \pm 0.007$ & $1.193 \pm 0.003$ & $1.172 \pm 0.002$ \\
\bottomrule
\end{tabular}
\end{center}

\paragraph{High-dimensional SNR sweep.}
For the activation-level synthetic pipeline, we generated latents from the human correlation matrix, projected through a random Gaussian matrix $\mathbf{W} \in \mathbb{R}^{d \times 16}$ with $d \in \{896, 2048, 4096, 5120\}$, and added isotropic Gaussian noise at SNR $\in \{0.1, 0.3, 1.0, 3.0, 10.0\}$. Over 400 runs ($4 \times 5 \times 20$ seeds), recovered slope depends on SNR with minimal dependence on $d$ within our range:

\begin{center}\small
\begin{tabular}{cccccc}
\toprule
SNR & $d = 896$ & $d = 2048$ & $d = 4096$ & $d = 5120$ \\
\midrule
$0.1$  & $1.39$ & $1.48$ & $1.54$ & $1.56$ \\
$0.3$  & $1.32$ & $1.30$ & $1.31$ & $1.31$ \\
$1.0$  & $1.22$ & $1.17$ & $1.14$ & $1.14$ \\
$3.0$  & $1.16$ & $1.10$ & $1.08$ & $1.07$ \\
$10.0$ & $1.11$ & $1.07$ & $1.04$ & $1.03$ \\
\bottomrule
\end{tabular}
\end{center}

Mantel $r$ between recovered and human matrices is consistently $\geq 0.97$ for all cells, confirming that the pipeline recovers correlation structure faithfully regardless of magnitude inflation.

\paragraph{Per-model adjustment.}
Pipeline baselines for each of the 9 cross-persona models, obtained from the synthetic grid using each model's hidden dimension and SNR proxy ($\rho / (1 - \rho)$), range from $1.11$ (Qwen~14B Instruct) to $1.48$ (Qwen~3B Instruct). Adjusted slopes (observed $-$ baseline) range from $-0.09$ (Qwen~3B Instruct) to $+0.66$ (Qwen~14B Instruct); mean adjusted slope is $+0.27$. Correlations of adjusted slope with behavioral slope ($r = 0.88$, $p = 0.002$) and log model size ($r = 0.68$, $p = 0.045$) remain substantial.

\paragraph{SNR proxy caveat.}
The SNR proxy used for per-model adjustment is derived from contrastive-direction reliability, not from the item-level ridge pipeline. A matched correction would require reliability estimates for the item-level predictions. The reported baseline and adjusted slopes are therefore approximate.

\section{Scale Details}
\label{app:scales}

Table~\ref{tab:scale_details} lists all 7 scales, their subscale composition, item numbers (1-indexed, matching the administered Chinese-language item order), and reverse-coded items. Response ranges vary by scale. All 114 items are administered in the Chinese translation used with our $N = 272$ participants. Subscale scores average item values after reverse coding where applicable.

The questionnaire files in the accompanying repository contain the exact Traditional Chinese item wording and response options administered to participants for all 114 items.

\begin{table}[t]
\centering\footnotesize
\caption{Complete subscale composition and reverse-coded items per scale.}
\label{tab:scale_details}
\begin{tabular}{llcl}
\toprule
Scale (range) & Subscale & Items (1-indexed) & Reverse-coded (within scale) \\
\midrule
IRI (1--5) & Perspective Taking & 3, 8, 11, 15, 21, 25, 28 & 3, 4, 7, 12, 13, 14, 15, 18, 19 \\
           & Fantasy & 1, 5, 7, 12, 16, 23, 26 & \\
           & Empathic Concern & 2, 4, 9, 14, 18, 20, 22 & \\
           & Personal Distress & 6, 10, 13, 17, 19, 24, 27 & \\
\midrule
PANAS (1--5) & Positive Affect & 1, 3, 5, 9, 10, 12, 14, 16, 17, 19 & (none) \\
             & Negative Affect & 2, 4, 6, 7, 8, 11, 13, 15, 18, 20 & \\
\midrule
POM (1--6) & Peace of Mind & 1--7 (all) & 5, 7 \\
\midrule
BFI-10 (1--7) & Extraversion & 1, 6 & 2, 6, 8, 9, 10 \\
              & Agreeableness & 2, 7 & \\
              & Conscientiousness & 3, 8 & \\
              & Neuroticism & 4, 9 & \\
              & Openness & 5, 10 & \\
\midrule
SCS (1--7) & Independent self & 1, 6, 7, 9, 11, 12, 15, 17, 20, 21, 22, 23 & (none) \\
           & Interdependent self & 2, 3, 4, 5, 8, 10, 13, 14, 16, 18, 19, 24 & \\
\midrule
SWLS (1--7) & Life Satisfaction & 1--5 (all) & (none) \\
\midrule
UCLA (1--4) & Loneliness & 1--20 (all) & 1, 5, 6, 9, 10, 15, 16, 19, 20 \\
\bottomrule
\end{tabular}
\end{table}

\section{Cross-Persona Details}
\label{app:cross_persona}

\paragraph{Activation extraction.}
For the single-scale activation probing paradigm (\S\ref{sec:amplification}), we hook the post-MLP residual-stream output of each transformer layer (i.e., the output of \texttt{model.model.layers[i]}, equivalent to the input to layer $i{+}1$) and record the activation at the \emph{last token position} of the prompt under each (subject, rotation, target-item) combination. All prompts use raw-text concatenation (persona system prompt $\|$ target-item question $\|$ ``\texttt{Answer:~}'')\footnote{The trailing space after the colon is intentional. The BPE tokenizers used by Llama and Qwen treat the response digit as its own token only if separated from the colon by whitespace; without the trailing space, the colon and digit can fuse into a single token, which would invalidate the option-likelihood scoring described in \S\ref{sec:setup}.} rather than the model's chat template, matching the behavioral pipeline and ensuring comparability between base and instruct variants. A candidate layer set is recorded per model (scaled with depth: layers 12--24 for Llama~8B; analogous mid-to-late ranges for other sizes); the ``best layer'' in Table~\ref{tab:cross_persona} is selected post-hoc from a per-model layer-wise sweep on the plateau region of high Mantel $r$ and slope, with ties broken in favour of higher Mantel $r$. Appendix~\ref{app:robustness} reports layer-wise sensitivity.

\paragraph{Directional asymmetry.}
The LLM cross-scale correlation matrix is not symmetric: cell $(i,j)$ and $(j,i)$ come from different persona rotations (Scale~$i$ as persona vs.\ Scale~$j$ as persona). We correlate the 21 upper-triangle entries of the activation-predicted and behavioral directed scale matrices, retaining each entry's source-to-target orientation. On Llama~8B Instruct, this correlation is $r = 0.94$ ($p < 0.0001$); on Qwen~14B Base, it is $r = 0.84$ ($p < 0.0001$). Thus the directed scale-pair pattern in activation-probe predictions is associated with that in behavioral output; this does not establish that the measured directions cause that behavior.

\section{Additional Behavioral Results}
\label{app:behavioral}

\paragraph{Cronbach's alpha for attenuation correction.}
The attenuation-corrected slopes reported in \S\ref{sec:setup} are computed by disattenuating the human correlation matrix only (unilateral correction; the LLM matrix is left untouched, as the LLM has no measurement-error reliability analogue at the subject level). Cronbach's $\alpha$ is estimated directly from the $N = 272$ Taiwanese participant responses using the standard formula $\alpha = \frac{k}{k - 1}\left(1 - \frac{\sum_i \sigma^2_i}{\sigma^2_T}\right)$, where $k$ is the number of items, $\sigma^2_i$ is the variance of item $i$, and $\sigma^2_T$ is the variance of the total score. Disattenuated correlations are $r_{\text{true}}(x, y) = r_{\text{obs}}(x, y) / \sqrt{\alpha_x \alpha_y}$; values that cross $\pm 1$ after disattenuation are clipped. For the seven-scale analysis, we use scale-level reliability estimates: one Cronbach's $\alpha$ is computed for each of the seven scales from all items in that scale (e.g., 28 items for IRI, 20 for PANAS, 7 for POM). The human scale-level correlation matrix is then disattenuated unilaterally using these seven scale reliabilities. These corrected scale-level slopes are reported separately from Figure~\ref{fig:behavioral}.

\paragraph{Subscale-level sensitivity to reliability cut.}
Computing the correction at the subscale level (16 subscales) instead of the scale level exposes a sensitivity to short-instrument reliability. Subscale-level $\alpha$ values range from $0.30$ (BFI Agreeableness, $k = 2$) to $0.92$ (POM Peace of Mind, $k = 7$); four of the five two-item BFI-10 subscales (Agreeableness, Conscientiousness, Neuroticism, Openness) yield $\alpha \in [0.30, 0.60]$, well below the conventional reliability bar. Cronbach's $\alpha$ is known to be unstable when computed on $k = 2$ items \citep{cortina1993coefficient}; BFI-10 was designed to use test-retest rather than internal-consistency reliability for this reason \citep{rammstedt2007measuring}. Disattenuating with these underestimated $\alpha$ values inflates the human correlations more than warranted, suppressing the corrected LLM-vs-human slope. Sweeping a minimum-$\alpha$ threshold and recomputing corrected slopes for $\geq 7$B models gives Table~\ref{tab:attenuation_subscale_sensitivity}: as more low-$\alpha$ subscales are excluded, the corrected slope generally rises and approaches the scale-level estimate. Restricting to $\alpha \geq 0.80$ subscales (the cut at which Cronbach's estimates are themselves reliable) yields corrected slopes of $1.22$--$1.36$, compared with the scale-level range of $1.19$--$1.39$. Both restrictions (scale-level aggregation and the high-$\alpha$ subscale subset) yield slopes above $1$, supporting the \S\ref{sec:setup} claim that amplification persists after disattenuation, while making explicit that the conclusion depends on having reliability estimates that are themselves trustworthy.
 
\begin{table}[t]
\centering\footnotesize
\caption{Subscale-level disattenuated slope for $\geq 7$B models as a function of the minimum-reliability cut. As more low-$\alpha$ subscales are excluded, the corrected slope generally rises and approaches the scale-level estimate.}
\label{tab:attenuation_subscale_sensitivity}
\begin{tabular}{lcccc}
\toprule
Cut & $n_{\text{subscales}}$ & $n_{\text{pairs}}$ & $\geq 7$B range & $\geq 7$B mean \\
\midrule
All 16 subscales       & 16 & 116--120 & $[0.62, 0.92]$ & 0.79 \\
$\alpha \geq 0.60$     & 12 &  62--66 & $[0.78, 1.05]$ & 0.92 \\
$\alpha \geq 0.70$     &  9 &  36 & $[0.80, 1.05]$ & 0.94 \\
$\alpha \geq 0.80$     &  5 &  10 & $[1.22, 1.36]$ & 1.30 \\
\midrule
Seven-scale matrix & 7 & 21 & $[1.19, 1.39]$ & 1.26 \\
\bottomrule
\end{tabular}
\end{table}
 
\begin{figure}[t]
\centering
\includegraphics[width=\linewidth]{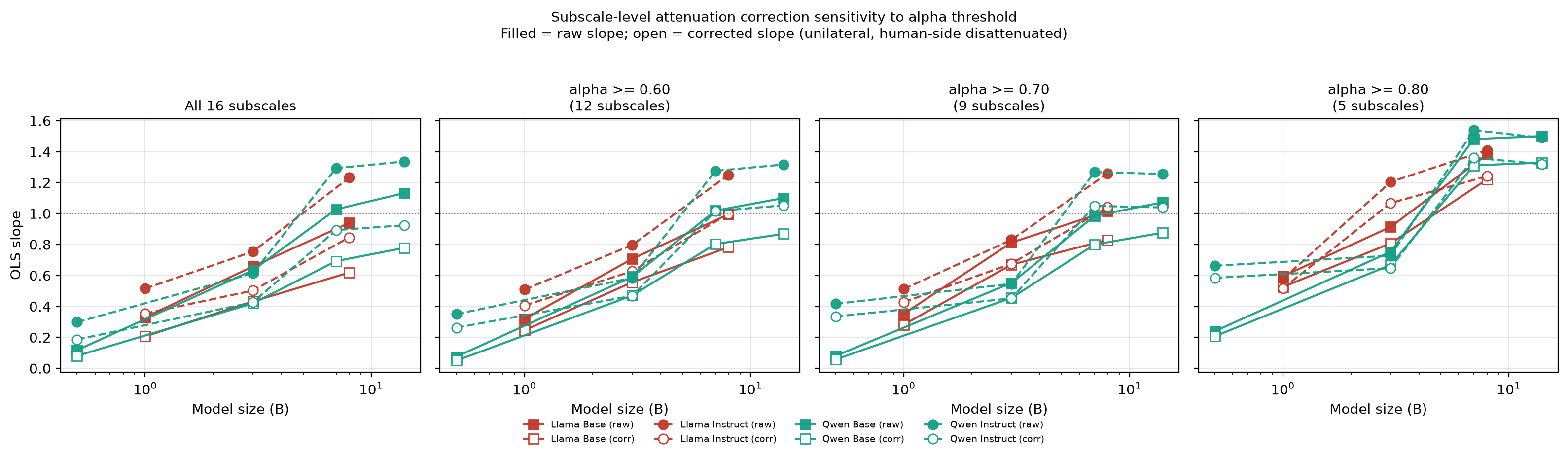}
\caption{Subscale-level attenuation correction sensitivity to the minimum-$\alpha$ cut. Filled markers: raw slope; open markers: corrected slope (unilateral, human-side disattenuated). At the most permissive cut (all 16 subscales) the four BFI-10 short subscales pull the corrected slope downward; restricting to $\alpha \geq 0.80$ removes them and gives corrected slopes of $1.22$--$1.36$ for the tested $\geq 7$B variants.}
\label{fig:attenuation_subscale_sensitivity}
\end{figure}

\paragraph{Per-scale alignment.}
Figure~\ref{fig:behavioral} reports mean behavioral alignment across seven persona-scale rotations. Per-scale numbers are not transcribed here because the pattern does not materially change the main conclusions.

\begin{figure}[t]
\centering
\includegraphics[width=0.6\linewidth]{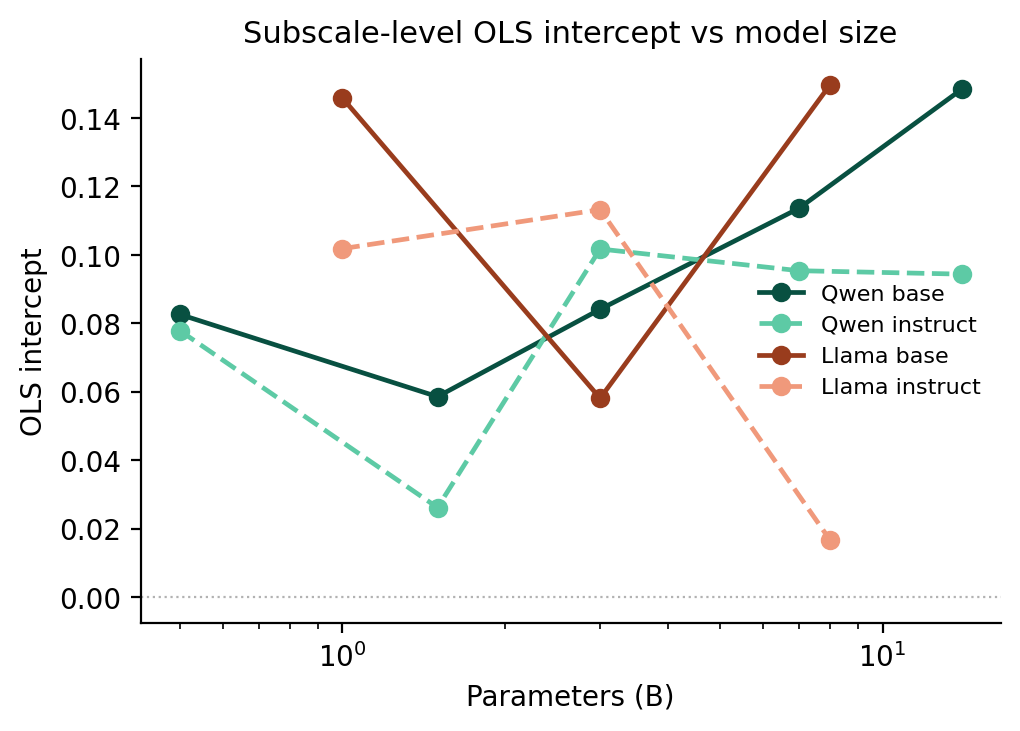}
\caption{OLS intercept of LLM-vs-human subscale-level cross-scale correlation regression as a function of model size. Colour indicates family (red: Llama, green: Qwen) and base/instruct as marked in the legend. The intercept does not exhibit a monotone scaling trend, is predominantly positive across models, and indicates a baseline offset in LLM-predicted correlations that is separable from the slope-based amplification reported in \S\ref{sec:setup}.}
\label{fig:intercept_scaling}
\end{figure}

\paragraph{Intercept inflation as a separable distortion.}
The amplification analyses in \S\ref{sec:setup} characterise LLM cross-scale correlations through their slope when regressed onto human correlations. OLS, however, has a second free parameter, the intercept, which captures a baseline offset: the predicted LLM correlation at $r_\text{human} = 0$. We report this subscale-level intercept across all 14 variants in Figure~\ref{fig:intercept_scaling}. Two patterns are notable. First, the intercept does not exhibit a clean scaling trend: it does not decrease with model size for either family. Second, intercepts are systematically positive across models (median across the 14 variants: $0.10$), and the two base models for which we computed split-half baselines (App~\ref{app:robustness}) lie beyond the right tail of the human split-half intercept distribution. LLMs therefore appear to introduce a positive baseline shift on top of the slope-based amplification documented in the main text: even subscale pairs that are uncorrelated in the human data tend to be predicted with mildly positive correlations.

We do not interpret this as a separate causal mechanism on the basis of the present data. Possibilities include a generic acquiescence-style response style that is incompletely cancelled by reverse-coding, a mild positive prior over correlations introduced by the in-context persona format, or residual influence of token-level co-occurrence in the prompt. Distinguishing these would require a controlled probe (e.g., correlations between truly orthogonal constructs in the prompt, or randomised-persona controls) that we did not run. We report the intercept here so that readers interpreting the slope numbers as a complete summary of LLM-vs-human distortion are aware that a non-trivial intercept component coexists with the slope effect and is not absorbed by it.

\FloatBarrier
\paragraph{Behavioral semantic similarity control.}
A potential concern for the representational-geometry results (App~\ref{app:geometry}) is that the apparent structure might be inherited from surface lexical similarity between item wordings. The same concern applies in principle to the behavioral pipeline; if LLMs match human structure, perhaps they are matching item-word similarity instead. We rerun the same control here: compute item-text embedding similarity, average within subscale to a $16 \times 16$ semantic similarity matrix $S$, and partial it out of the behavioral $r(B, H)$ alignment. We report two embedding sources for robustness: (a) each model's own input embedding layer (mean-pooled across item tokens), the same source used in App~\ref{app:geometry}; (b) the multilingual-e5-large sentence encoder, an external semantic encoder shared across all 14 variants. Across 14 variants, both sources yield mean partial correlations very close to the zero-order alignment (mean drop $0.002$ for source (a), $0.008$ for source (b); maximum absolute drop across the 14 variants is $0.013$ for (a) and $0.036$ for (b)). These two measured semantic-similarity controls do not materially change alignment; they do not rule out broader learned psychological associations.
 
\begin{figure}[t]
\centering
\includegraphics[width=\linewidth]{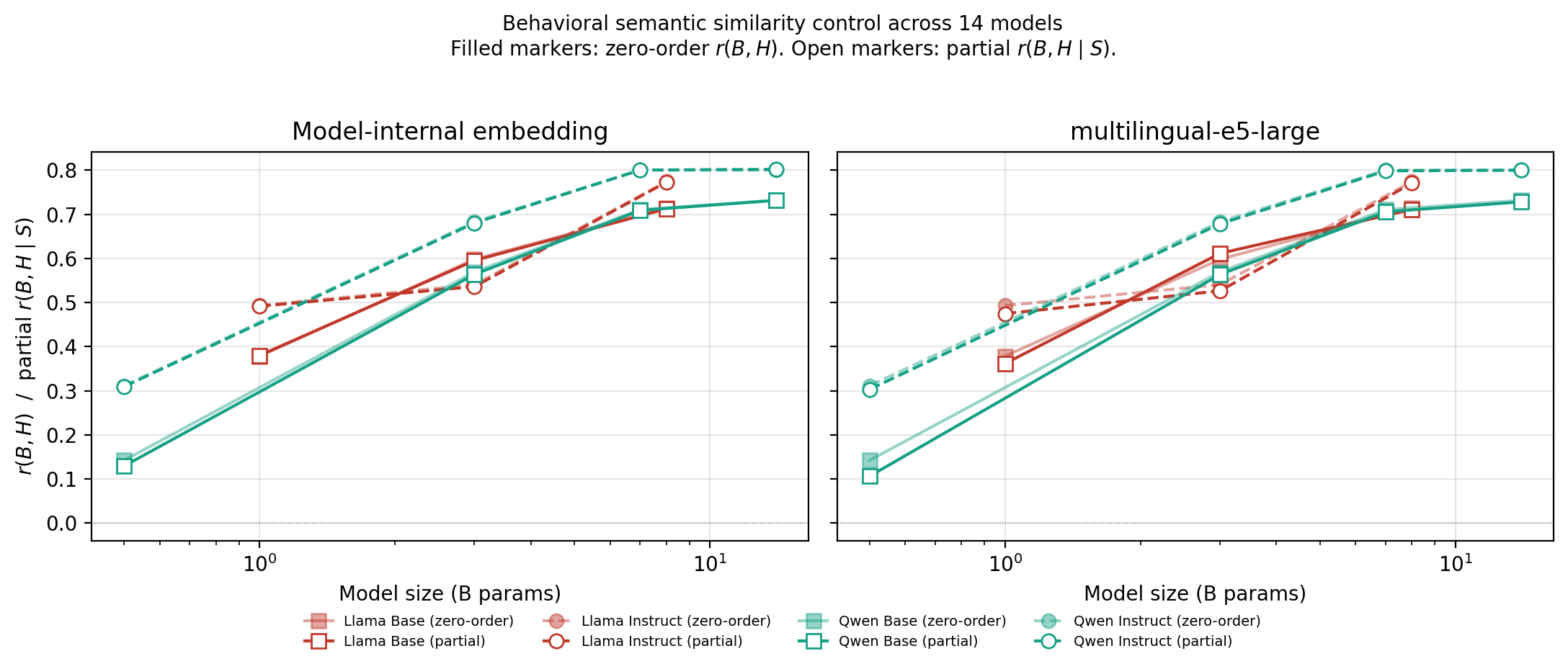}
\caption{Behavioral semantic similarity control across 14 variants. Filled markers: zero-order $r(B, H)$. Open markers: partial $r(B, H \mid S)$ controlling for item embedding similarity $S$. Left: each model's input embedding layer; right: multilingual-e5-large. Partial and zero-order values are visually indistinguishable for all variants in both panels.}
\label{fig:behavioral_semantic_control}
\end{figure}

\section{Representational Geometry Details}
\label{app:geometry}

\begin{figure}[t]
\centering
\includegraphics[width=\linewidth]{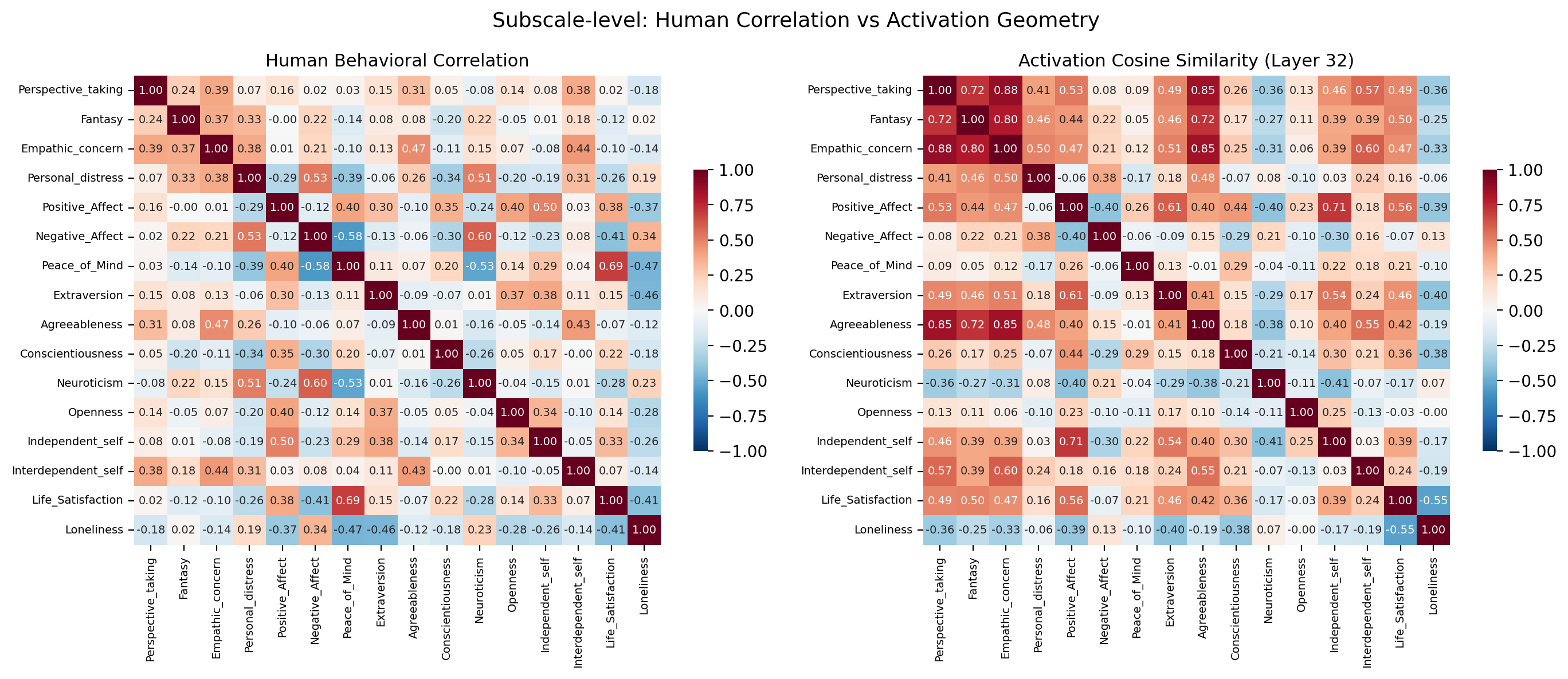}
\caption{Human subscale correlation matrix (left) vs.\ Qwen~14B Instruct activation cosine similarity matrix (right, layer~32). Mantel $r = 0.62$ at this layer ($0.66$ at selected layer~36). Visible block structure corresponds to within-scale subscale groupings (e.g., IRI PT/FS/EC/PD; Big Five E/A/C/N/O) and cross-scale relationships (e.g., positive PA--LifeSat, positive NA--Loneliness). The corresponding Qwen~14B Base matrix (Mantel $r = 0.13$) shows no comparable block structure.}
\label{fig:qwen14b_heatmap}
\end{figure}

\paragraph{Qwen~14B Instruct geometry: visual comparison.}
Figure~\ref{fig:qwen14b_heatmap} shows the layer~32 matrices (Mantel $r = 0.62$; the selected layer~36 gives $r = 0.66$). The human subscale correlation matrix (left) and Qwen~14B Instruct's activation cosine similarity matrix (right) show matching block structure: within-scale subscale groupings (e.g., IRI PT/FS/EC/PD; Big Five E/A/C/N/O) cluster in both, and cross-scale relationships such as positive PA--LifeSat and positive NA--Loneliness appear in both. Qwen~14B Base, which shares pretraining but differs only in instruction tuning, shows Mantel $r = 0.13$ and no comparable block structure (matrix omitted).

\begin{figure}[t]
\centering
\includegraphics[width=\linewidth]{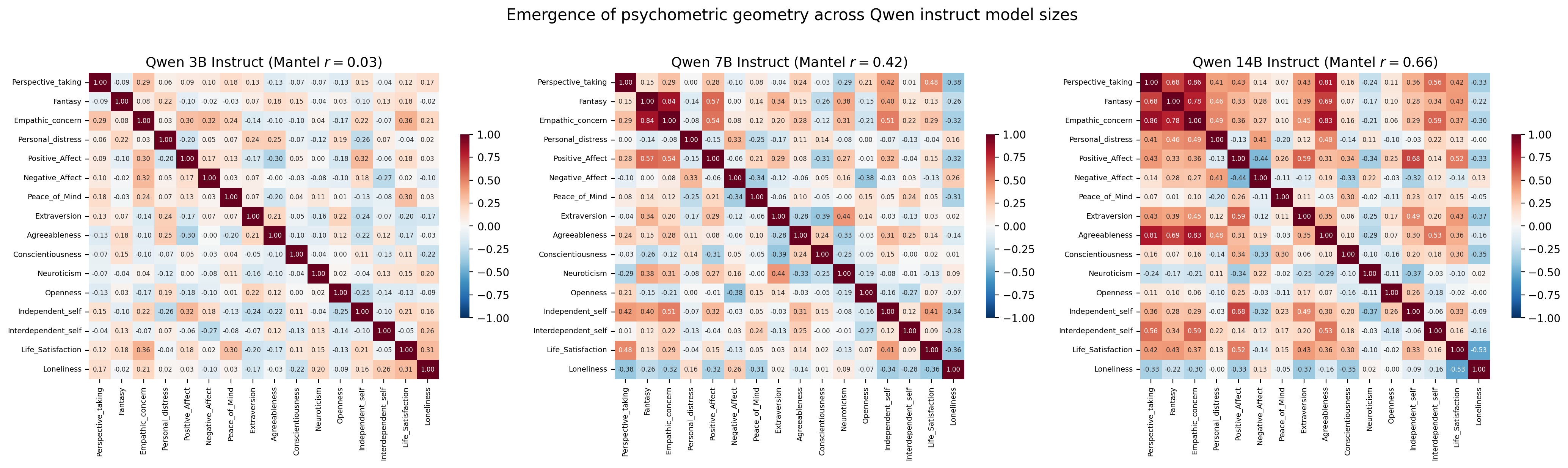}
\caption{Emergence of psychometric geometry across Qwen instruct model sizes. Activation cosine similarity matrices for Qwen~3B Instruct (left, Mantel $r = 0.03$), Qwen~7B Instruct (middle, $r = 0.42$), and Qwen~14B Instruct (right, $r = 0.66$). Matrices use identical subscale ordering and colour scale. Block structure corresponding to human psychometric correlations becomes increasingly visible with scale.}
\label{fig:emergence_grid}
\end{figure}

\paragraph{Emergence of geometry with scale.}
Figure~\ref{fig:emergence_grid} visualises the scaling curve reported in Figure~\ref{fig:representational}. Qwen~3B Instruct (Mantel $r = 0.03$) shows near-random pattern; Qwen~7B Instruct ($r = 0.42$) shows partial block structure; Qwen~14B Instruct ($r = 0.66$) shows clean alignment with the human correlation matrix. Geometry emergence is continuous rather than a sharp threshold, but the rate of emergence is rapid between 7B and 14B.

\paragraph{Contrastive direction extraction procedure.}
For each of the 16 subscales, we construct paired \emph{high} and \emph{low} conditions. In the high condition, the target subscale's items (e.g., all 7 IRI Perspective Taking items) are set to the maximum value of the scale's response range (e.g., 5 for IRI); in the low condition, to the minimum. All other items across all 7 scales (to make a 114-item mega-prompt) are set to randomly sampled values from the scale range, independently per replication. The resulting system prompt is concatenated with a probe question (see below); the model generates up to 128 continuation tokens, and we record the mean residual-stream activation at each target layer over the generated token positions. We use $N = 10$ random background replications per (subscale, condition) pair and $N = 20$ probe questions per condition, yielding $16 \times 2 \times 10 \times 20 = 6{,}400$ condition--probe generations per model. The direction $\mathbf{d}_k$ for subscale $k$ is the difference of means between the $200$ high-condition activations and the $200$ low-condition activations.

\paragraph{Probe questions.}
The 20 open-ended probe questions used for contrastive extraction are in Traditional Chinese. They are not questionnaire items. The full bank contains 40 questions organized into eight broad domains: daily life, social interaction, self-reflection, stress and emotion, future goals, values, concrete situations, and interests and preferences. This analysis uses the first 20 questions. Examples ask how a person spends a weekend, makes an important decision, or maintains a friendship. The probes elicit continuations under the high and low conditions. The exact questions and first-20 selection are included in the accompanying repository.

\paragraph{Layer sweep.}
We record activations at candidate layers spanning mid-to-late transformer depth, scaled to each model's number of layers: for Llama~8B ($L = 32$) we use layers \{12, 14, 16, 18, 20, 22, 24\}; for Qwen~14B ($L = 48$), \{16, 20, 24, 28, 32, 36\}. Layer~16 (${\approx}50\%$ depth) is our default for Llama~8B, matching both the CAA literature and our steering effects. For the contrastive-direction analysis, the selected layer is chosen post-hoc by the geometry metric. A geometry-specific adjacent-layer check is in App~\ref{app:robustness}.

\paragraph{Mantel test.}
For the main contrastive geometry and steering specificity analyses, we use the Mantel test with 10{,}000 random permutations of one matrix's row/column indices to generate a null distribution for the off-diagonal correlation. Double-centered Mantel (used for steering specificity) first applies $\mathbf{M}_{\text{dc}} = \mathbf{M} - \mathbf{r}_{\text{row}} - \mathbf{c}_{\text{col}} + \mu$, removing generic row/column shifts before permutation.

\paragraph{Semantic similarity control.}
To test whether direction geometry reflects item-wording similarity, we compute an item-embedding similarity matrix and partial it out of the alignment between contrastive direction geometry and human psychometric structure. We report two embedding sources to ensure the control is not specific to either choice: (a) each model's own input embedding layer (mean-pooled across item tokens), which captures the model's own representation of item lexical content; (b) the multilingual-e5-large sentence encoder \citep{wang2024multilingual}, an external general-purpose semantic encoder shared across all 14 variants. For each source we aggregate to a $16 \times 16$ subscale-level semantic similarity matrix $S$ by mean-pooling item embeddings within each subscale. Across 14 variants the mean zero-order $r(D, H)$ between contrastive direction geometry and the human psychometric matrix is $0.115$; the mean partial $r(D, H \mid S)$ is $0.113$ with embedding source (a) and $0.110$ with source (b). The mean drop from zero-order to partial is below $0.01$ for both sources, with maximum absolute drop across the 14 variants below $0.03$. On Llama 8B Instruct specifically, $r(D, H) = 0.368$ and the partial correlation is higher by $0.007$ (source (a)) or $0.020$ (source (b)). These two measured semantic-similarity matrices account for little of the observed alignment; they do not test whether the models use broader learned psychological priors. The same control is applied to the behavioral pipeline in App~\ref{app:behavioral} (Figure~\ref{fig:behavioral_semantic_control}); the conclusion is the same.
 
\begin{figure}[t]
\centering
\includegraphics[width=\linewidth]{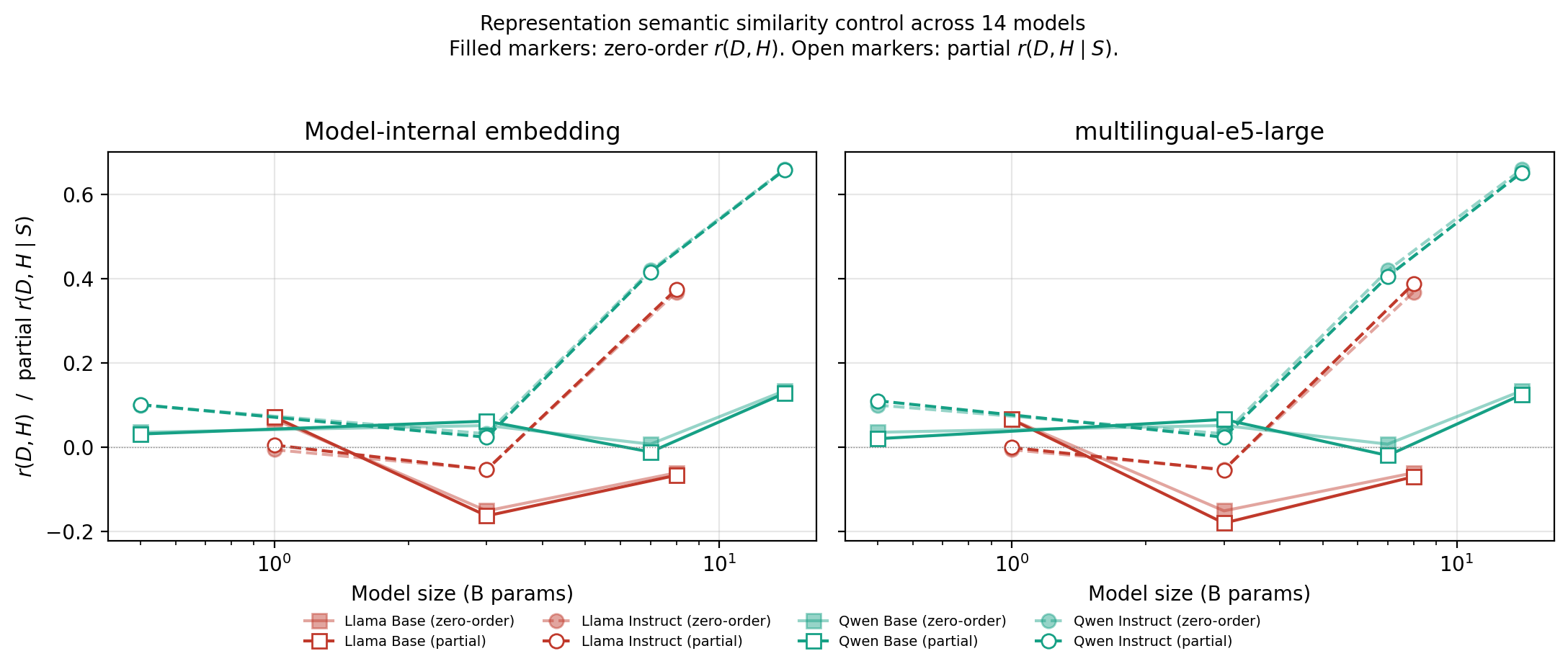}
\caption{Representational semantic similarity control across 14 variants. Filled markers: zero-order $r(D, H)$ between contrastive direction cosine similarity and the human subscale correlation matrix. Open markers: partial $r(D, H \mid S)$ controlling for item embedding similarity $S$. Left: each model's input embedding layer; right: multilingual-e5-large.}
\label{fig:representational_semantic_control}
\end{figure}

\section{Steering Details}
\label{app:steering}

\paragraph{Regression direction extraction.}
Steering directions differ from the contrastive directions used for geometry analysis: we fit ridge regression per subscale, mapping $N = 272$ subjects' layer-$L$ activations to their observed subscale scores. Ridge regularization is selected by 5-fold cross-validation over $\alpha \in \{0.1, 1, 10, 100, 1000\}$ (via \texttt{RidgeCV}). The direction for subscale $k$ is the normalized weight vector $\mathbf{d}_k = \mathbf{w}_k / \|\mathbf{w}_k\|_2$. Cross-validated prediction $r$ for each subscale provides a quality check; on Llama~8B Instruct, 13 of 16 subscales achieve CV $r > 0.2$.

\paragraph{Activation steering.}
We register a forward hook on the residual-stream output of layer $L$ (\texttt{model.model.layers[L]}) that adds $\alpha \cdot \mathbf{d}_k$ to the residual at the \emph{last token position} of the prompt (the position that immediately precedes generation), using the unit-normalised direction $\mathbf{d}_k$ defined above. We sweep $\alpha \in \{-2, -1, -0.5, 0, 0.5, 1, 2\}$ for each (source subscale, target item) pair and record the model's response log-probability distribution. Steering is applied at one token position in the model's native fp16 activation units; we do not rescale the direction by an empirical activation magnitude.

\paragraph{EV metric and steering-response matrix.}
For each target item with response options $\{1, 2, \ldots, K\}$, we compute the expected value $\text{EV} = \sum_{v=1}^{K} v \cdot p(v)$, which respects both the central tendency and the spread of the response distribution (unlike argmax). The steering effect is quantified per (source subscale $s$, target subscale $t$) as $\Delta\text{EV}_{s,t} = \text{slope of EV vs. } \alpha$, computed by OLS over the 7 $\alpha$ values. The regression-direction run uses all 16 source and 16 target subscales. A separate contrastive-direction run uses the 9 source subscales with split-half Spearman--Brown direction reliability $\geq 0.3$ and the same 16 targets. Specificity is assessed by computing Mantel $r$ between the double-centered steering-response matrix and the corresponding human psychometric correlation matrix, using the symmetrized square submatrix shared by source and target subscales (16 $\times$ 16 for regression directions; 9 $\times$ 9 for contrastive directions).

\paragraph{Per-model results.}
In the Llama~8B Instruct layer-16 regression-direction run, double-centered Mantel specificity is significant ($r = 0.28$, $p = 0.009$), as reported in the main text. The separate $9 \times 16$ contrastive-direction run gives $r = 0.14$, $p = 0.216$. Per-model steering runs show positive specificity in multiple instruct models, including strong double-centered effects in Llama~3B Instruct and Qwen~3B Instruct. The effects do not scale monotonically with model size, and a separate Llama~8B Instruct run is weaker than the layer-16 run reported above. Direction-estimation noise and the fixed steering magnitude are possible contributors, but these runs do not distinguish them. Formal layer-wise and magnitude sweeps remain future work.

\end{document}